%% file: sample-manuscript.tex
\documentclass[manuscript,screen, nonacm]{acmart}
\AtBeginDocument{%
  }

\setcopyright{acmlicensed}
\copyrightyear{2025}
\acmYear{2025}
\acmDOI{XXXXXXX.XXXXXXX}
\acmISBN{978-1-4503-XXXX-X/2018/06}

\usepackage{xcolor}         
\usepackage{booktabs}
\usepackage{longtable}
\usepackage{multirow}
\usepackage{rotating}

\definecolor{ForestGreen}{RGB}{34,139,34} 

\usepackage{tikz}
\usepackage{forest}
\usetikzlibrary{arrows.meta, positioning, calc}

\forestset{
  my tree/.style={
    for tree={
      grow'=0,
      draw,
      rounded corners=8pt,
      thick,
      align=left,
      font=\small,
      minimum height=8mm,
      inner xsep=5pt,
      inner ysep=3pt,
      l sep=8mm,
      s sep=4mm,
      edge={gray!70, thick},
      parent anchor=east,
      child anchor=west,
      anchor=west,
      edge path={
        \noexpand\path[draw,\forestoption{edge}]
        (!u.east) -- +(6pt,0) |- (.west)\forestoption{edge label};
      },
    }
  },
  redbox/.style={draw=red!70, fill=red!8},
  greenbox/.style={draw=green!60!black, fill=green!10},
  bluebox/.style={draw=cyan!70!blue, fill=cyan!10},
  rootbox/.style={
    draw=gray!60,
    fill=gray!10,
    minimum width=14mm,   
    minimum height=55mm,  
    align=center
  }
}

\begin{document}

\title{A Survey of AI-Generated Video Evaluation}

\author{Xiao Liu}
\authornote{Equal Contribution.}
\email{xioliu@ucdavis.edu}
\author{Xinhao Xiang}
\authornotemark[1]
\email{xhxiang@ucdavis.edu}
\author{Zizhong Li}
\authornotemark[1]
\email{zzoli@ucdavis.edu}
\affiliation{%
  \institution{IFM Lab, University of California, Davis}
  \country{USA}
}

\author{Yongheng Wang}
\affiliation{%
  \institution{University of California, Davis}
  \country{USA}}
\email{yhgwang@ucdavis.edu}

\author{Zhuoheng Li}
\affiliation{%
  \institution{IFM Lab, University of California, Davis}
  \country{USA}}
\email{pipli@ucdavis.edu}

\author{Zhuosheng Liu}
\affiliation{%
  \institution{IFM Lab, University of California, Davis}
  \country{USA}}
\email{zsliu@ucdavis.edu}

\author{Weidi Zhang}
\affiliation{%
  \institution{University of California, Davis}
  \country{USA}}
\email{wdizhang@ucdavis.edu}

\author{Weiqi Ye}
\affiliation{%
  \institution{University of California, Davis}
  \country{USA}}
\email{vikye@ucdavis.edu}

\author{Jiawei Zhang}
\authornote{Corresponding Author}
\affiliation{%
  \institution{IFM Lab, University of California, Davis}
  \country{USA}}
\email{jiwzhang@ucdavis.edu}

\renewcommand{\shortauthors}{Liu et al.}

\begin{abstract}

The rapid advancement of AI in video generation presents new challenges in evaluating AI-generated content. Unlike images or text, videos involve complex spatial and temporal dynamics. This survey identifies the emerging field of AI-Generated Video Evaluation (AIGVE), highlighting the importance of assessing how well AI-generated videos align with human perception and instructions. By outlining the strengths and gaps in current approaches, we advocate for the development of more robust and nuanced evaluation frameworks that can handle the complexities of video content, which include not only the conventional metric-based evaluations, but also the current human-involved evaluations, and the future model-centered evaluations.

\end{abstract}



\received{20 February 2007}
\received[revised]{12 March 2009}
\received[accepted]{5 June 2009}

\maketitle

\input{content/1-introduction}

\input{content/2-video_gen_model}

\input{content/3-prelim}

\input{content/4-vqa}

\input{content/5-tv_align}

\input{content/6-multi_asp}
\input{content/8-future}

\input{content/10-conclusion}

\bibliographystyle{ACM-Reference-Format}
\bibliography{sample-base,anthology, custom}

\appendix
\end{document}

%% file: content/1-introduction.tex
\section{Introduction}
\label{sec:intro}

With the introduction and widespread integration of large generative models like ChatGPT \cite{openai2023gpt4}, Sora \cite{sora2024}, LLaMA \cite{touvron_llama_2023}, and the recent Meta Movie Gen \cite{moviegen2024}, AI-generated content has become increasingly significant in both the production and consumption of contents. In the domain of production, text and video professionals increasingly use generative tools to create and enhance content, from scripts and articles to complex visual sequences, which would traditionally require extensive time and effort to achieve manually, thus streamlining creative workflows and enhancing productivity \cite{10.1257/jel.20231736, 10.1145/3633453, Albadarin2024, ADGVE, liu2024gpta}.
On the consumer front, reliance on outputs from generative models has also become commonplace, with applications ranging from information retrieval to the automation of routine tasks. This shift represents a significant transformation from the pre-2023 era, when such tasks were predominantly manual.

\begin{figure}[t!]
    \centering
    \includegraphics[width=0.75\linewidth]{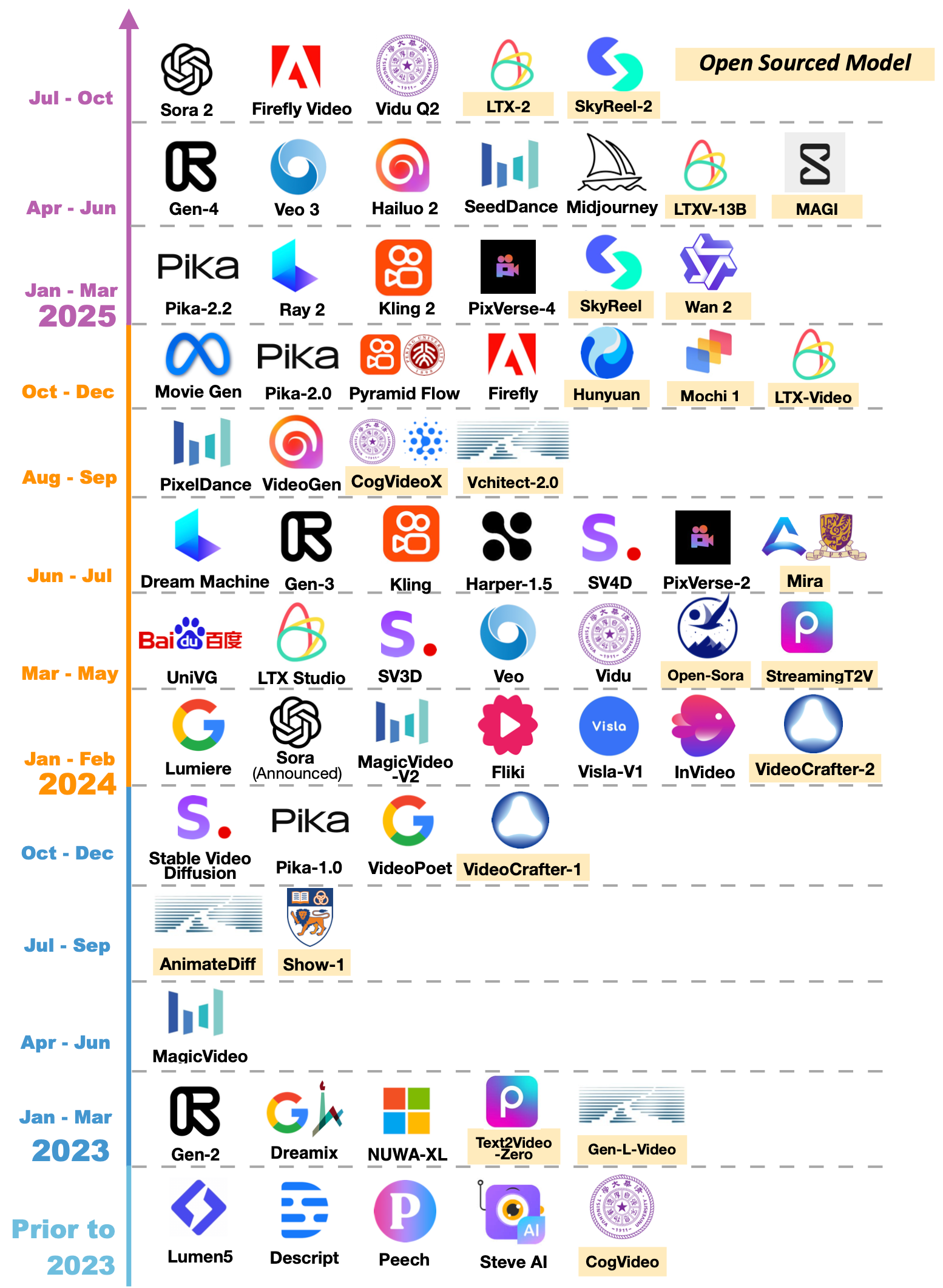}
    \caption{Evolution of Video Generation Models Over Time.}.
    \label{fig:video_geneation_model_timeline}
    \vspace{-5mm}
\end{figure}

As this trend continues to evolve, the methods that can automatically evaluate these AI-generated contents become crucial. These methods help ensure that such content aligns well with human perceptions and instructions. While the evaluation of AI-generated text and images has been thoroughly studied, using techniques including word and pixel matching \cite{lin2004rouge, heusel2018ganstrainedtimescaleupdate, salimans2016improvedtechniquestraininggans, Papineni2002BleuAM}, sophisticated modeling \cite{hessel2022clipscore, zhang2020bertscore, guan2024qmamba}, and Large/Vision Language Model evaluation \cite{fu2023gptscore, ku2023viescore, hu2023tifa, zhang2023gpt}, the assessment of AI-generated videos presents unique challenges that are yet to be comprehensively addressed. This discrepancy highlights a significant gap in research, particularly at a time when advanced video generation models are rapidly evolving, as illustrated in Figure \ref{fig:video_geneation_model_timeline}, and video content is gaining increasing prominence in both professional and personal domains.

Video, by its nature, incorporates both spatial complexity and temporal dynamics, making its evaluation intrinsically more complex than static images or text. Traditional Video Quality Assessment (VQA) metrics focus on technical aspects such as compression effects, transmission quality \cite{10.1145/3240508.3240643, 7459167, 7442861, PENG20171, 1588966}. Recent research has shifted focus towards evaluating the perceptual quality of user-generated, in-the-wild videos, considering factors like blur, motion stability, and noise levels \cite{wu_exploring_2023-1, wang2019youtube, 10.1145/3581783.3611860}. These methods assess whether a video effectively conveys visual information to the viewer. However, in the era of AI-Generated Content (AIGC), an additional critical aspect is whether the generated content aligns with the creator's instruction, which remains less explored within current VQA frameworks.

Besides, research in the evaluation of AI-generated videos is rapidly emerging but remains unstructured. The continuous introduction of new models and evaluation metrics complicates identifying comprehensive resources, as they are often scattered across various domains. Moreover, those methods focus on various evaluation aspects, which can overlap or be entirely disjoint. As a result, the absence of a unified framework impedes progress, resulting in fragmented research efforts. Therefore, there is a critical need for a more cohesive approach in this fast-moving area.

In this survey, we aim to highlight a new area focused on AI-Generated Video Evaluation (AIGVE). To devote our effort, we have collected and integrated existing research related to this field to help academic researchers and industrial practitioners locate the essential foundational knowledge. Our focus is on existing works from related research areas, and we have conducted extensive research on fields such as VQA \cite{min_perceptual_2024}, multimodal text-visual alignment \cite{yarom_what_nodate}, and recent emerging AIGVE evaluation methods \cite{liu2023evalcrafter, huang_vbench_2023, miao2024t2vsafety}. By exploring and categorizing methods related to AIGVE, we aim to build a solid foundation for AIGVE and support future research efforts in this rapidly evolving area.

This survey is organized as follows. Sections \ref{sec:intro} and \ref{sec:video_generators} present the motivation behind studying AI-Generated Video Evaluation, outlining both the rapid development of video generation technologies and summarizing their error types, introducing the emerging need for systematic evaluation frameworks. Section \ref{sec:aigve} reviews the current progress in AIGVE, summarizing benchmark datasets, evaluation paradigms, and methodological advances that shape the present research landscape. Building on this foundation, Sections \ref{sec_preception} and \ref{sec_instruction} examine two critical branches of the field: alignment with human perception and alignment with human instructions, respectively, providing categorized analyses of representative datasets, metrics, and models in each direction. Finally, Section \ref{sec_future} discusses future research opportunities and open challenges, highlighting promising directions for advancing evaluation methodologies in the evolving AIGVE ecosystem.


Our contributions through this survey are summarized as:

\begin{itemize}

    \item \textbf{Highlighting an Emerging Field:} We propose and emphasize the need for a new research area on AI-Generated Video Evaluation (AIGVE).

    \item \textbf{Comprehensive Review of Existing Evaluation Methods:} This survey provides a systematic and comprehensive review of current methodologies relevant to AIGVE from multiple research fields. We categorize and analyze these approaches to provide a well-structured outline of the existing landscape.

    \item \textbf{Guidance for Future Research Directions:} We also locate several potential areas that call for more future investigations and development in AIGVE. These areas include integrating evaluation frameworks with vision language models, enhancing the interpretability of evaluation scores, and addressing the ethical and safety considerations of these frameworks. This survey aims to serve as a foundational resource for researchers and industrial practitioners, providing insights that can guide the advancement of more effective and comprehensive evaluation methodologies for AI-generated video content.


\end{itemize}

%% file: content/2-video_gen_model.tex
\begin{figure}[!t]
    \centering
    \includegraphics[width=0.80\linewidth]{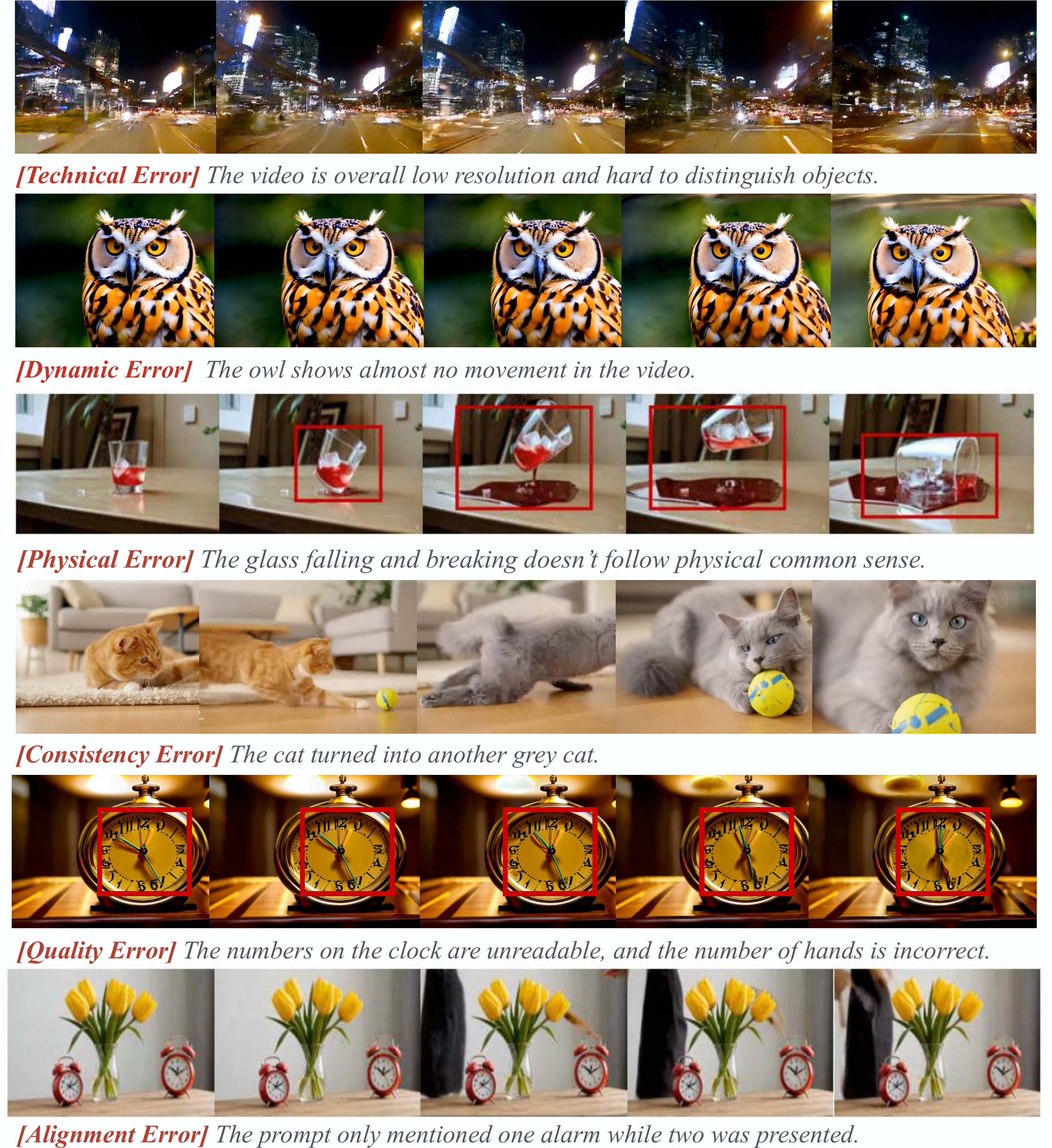}
    \caption{Illustration of common errors in AI-generated videos. We summarize six representative failure modes frequently observed in generated content: (1) technical errors, where videos exhibit low resolution or severe blur; (2) dynamic errors, where little or no meaningful motion occurs across frames; (3) physical errors, where object behaviors violate physical common sense; (4) consistency errors, where object identities or appearances change unexpectedly over time; (5) quality errors, where objects contain distorted structures or incorrect details; and (6) alignment errors, where the generated content deviates from the given textual instructions.}
    \label{fig:01}
\end{figure}
\section{Common Error Types in AI-Generated Videos}
\label{sec:video_generators}


Generating videos that align with real-world dynamics and physical laws has become a central goal in recent research. With the rapid development of generative modeling techniques, video generation systems have achieved improvements in visual fidelity, temporal coherence, and controllability. Existing approaches can generally be categorized into three major paradigms: Generative Adversarial Networks, Autoregressive Transformer models, and Diffusion Models \cite{kumar2025bridgingtextvideogeneration}.

The rapid development of these generative paradigms has substantially improved the capabilities of modern video generation systems. Both open-source and commercial models can now produce videos with increasingly higher fidelity, longer durations, and stronger multimodal conditioning. Despite these advancements, ensuring the reliability and realism of generated videos remains challenging. Generated content may still exhibit perceptual inconsistencies or deviate from the intended instructions. Broadly speaking, these limitations can be grouped into two major categories: (1) violations of human perceptual expectations and real-world dynamics, and (2) misalignment between generated content and the creator's instructions.

To better characterize these limitations, we summarize six common types of errors frequently observed in AI-generated videos, as illustrated in Figure~\ref{fig:01}. These error categories highlight key challenges in achieving perceptually realistic, temporally coherent, and instruction-aligned video synthesis.

\textbf{Technical Error.}
Technical errors arise from limitations in the generation pipeline that degrade the overall visual quality of the produced video. These include low spatial resolution, excessive blurring, compression artifacts, or low frame rates that result in unstable playback.

\textbf{Dynamic Error.}
Dynamic errors occur when the generated video fails to exhibit meaningful temporal motion. In such cases, the content remains nearly static, resembling a sequence of still images rather than a evolving video.

\textbf{Physical Error.}
Physical errors refer to violations of real-world physical principles or commonsense reasoning, such as unrealistic object motion, implausible collisions, or incorrect interactions with gravity.

\textbf{Consistency Error.}
Consistency errors occur when objects or scene elements change unexpectedly across frames, such as sudden alterations in color, shape, or identity, undermining 3D structural coherence of the depicted scene~\cite{TTSA3R}.

\textbf{Quality Error.}
Quality errors correspond to deficiencies in the structural fidelity of generated objects, including distorted geometry, incorrect shapes, missing components, or unreadable fine details.

\textbf{Alignment Error.}
Alignment errors arise when the generated video fails to follow the provided instructions or textual prompts, such as introducing unintended objects or omitting required actions.

Overall, these error categories highlight that, despite significant advances in generative modeling, current video generation systems continue to struggle with simultaneously maintaining visual fidelity, temporal coherence, physical plausibility, and semantic alignment with user instructions. Besides, these errors are often subtle and difficult to detect through casual inspection, making systematic evaluation particularly challenging. Although human assessment remains the most reliable approach for judging the realism of generated videos, comprehensive automated frameworks for consistently evaluating the quality and correctness of AI-generated videos are still open questions.

%% file: content/3-prelim.tex
\begin{figure*}[!ht]
    \centering
    \includegraphics[width=0.85\linewidth]{figures/AIGVE_mind.pdf}
    \caption{The development and overview of \textcolor[HTML]{7BAA82}{AI-Generated Video Evaluation (AIGVE)}. AIGVE was built on two initially separate aspects: 1) \textcolor[HTML]{D49A89}{Alignment with human perception} , and 2) \textcolor[HTML]{8A9AAE}{Alignment with human instructions}. Note that the timeline scales are different for two aspects. Release date represents the date that this survey is released.}
    \label{fig:timeline}
    \vspace{-10pt}
\end{figure*}

\section{AI-Generated Video Evaluation}
\label{sec:aigve}

The field of evaluating AI-generated videos is still in its early stages. As synthesized video content becomes increasingly prevalent, there is a growing demand for effective evaluation methods that align with both the intentions of creators and the perceptions of viewers. This survey aims to outline a preliminary framework for AI-Generated Video Evaluation (AIGVE), recognizing that the definitions and methodologies will continue to evolve as the field matures.

By extensively reviewing and categorizing existing research \cite{min_perceptual_2024, wu_exploring_2023-1, liu_mmbench_2024, liu2023evalcrafter}, we propose that AI-generated videos should ideally satisfy two complementary criteria: 1) alignment with human perception and 2) alignment with human instructions. These two dimensions also correspond to different categories of errors that current evaluation methods aim to identify and mitigate.

\textbf{Alignment with Human Perception}. This aspect focuses on evaluating video quality from a human perceptual perspective. It is primarily designed to identify and address a range of perceptual and generation-related errors, including Technical Error, Dynamic Error, Physical Error, Consistency Error, and Quality Error introduced in Section \ref{sec:video_generators}. 

Traditional evaluation metrics emphasize factors such as resolution, clarity, and the absence of noise \cite{min_perceptual_2024}. However, AI-generated videos introduce additional complexity beyond these low-level signals. Videos must exhibit coherent motion over time (dynamic consistency), maintain object and scene consistency across frames, and adhere to real-world physical constraints \cite{sanchez2020learning, tang2023intrinsic}. This includes realistic texture rendering, accurate color representation, and physically plausible interactions. Therefore, alignment with human perception extends beyond visual fidelity to encompass temporal coherence and physical realism, ensuring that generated videos are both visually convincing and perceptually believable.

\textbf{Alignment with Human Instructions}. With the advancement of generative AI technologies, a new challenge has emerged: ensuring that generated videos faithfully follow detailed human instructions, which are typically expressed in textual form. This aspect is primarily designed to identify and resolve Alignment Error in Section \ref{sec:video_generators}, where the generated content deviates from the intended semantics, actions, or narrative described by the user.

This involves generating content that accurately reflects described scenarios, objects, actions, and temporal sequences, thereby fulfilling the creative and communicative objectives conveyed by the creator. Alignment with human instructions ensures that the generated video is not only visually plausible but also semantically correct and faithful to the input prompt, making it a true realization of user intent.

Figure \ref{fig:timeline} illustrates the development and overview of both aspects and highlights the emergence of AIGVE. Initially, the evaluation of video alignment with human perception and instructions were separate research areas. With the rise of AI-generated videos, both areas need to be considered when evaluating AI-generated videos.

In the remainder of this section, we present the development of AIGVE. It begins by detailing the creation of the benchmark dataset, transforming video-opinion pairs into video-instruction-opinion triplets for more nuanced evaluation. The evaluation methods are then categorized into two main approaches: metric collection evaluation, which utilizes existing metrics to assess various aspects of video quality, and modeling evaluation, where new models are developed and trained on the collected datasets to simulate human judgment.

\subsection{Benchmark Datasets for AI-Generated Video Evaluation}

The construction of benchmark datasets is a crucial prerequisite for training evaluation models and for systematically assessing the performance of AI-generated video systems. Beyond dataset scale and annotation fidelity, an effective benchmark must be designed in accordance with the \emph{specific evaluation objective} it is intended to support. From a technical usage perspective, existing AIGVE datasets vary substantially in their emphasis on holistic perceptual quality, temporal dynamics, semantic reasoning, or real-world and safety-related constraints. Rather than treating these benchmarks as interchangeable, it is essential to understand their functional roles and appropriate application scenarios.

\begin{figure}[t]
    \centering
    \small
    \includegraphics[width=0.70\linewidth]{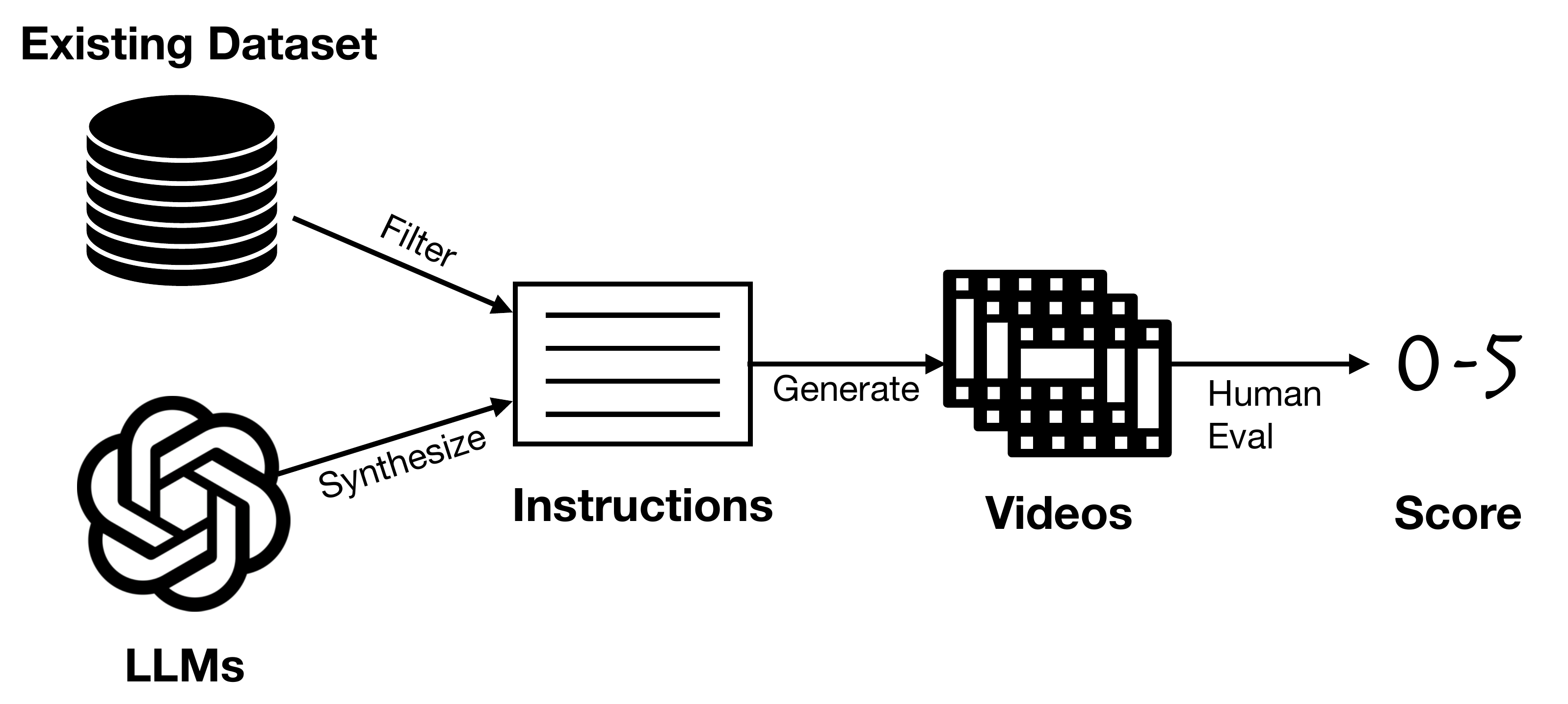}
    \caption{AIGVE Benchmark Dataset Collection Process.}
    \label{fig:data_collection}
\end{figure}

To better align evaluation with human perception and instruction-following behavior, recent benchmarks extend the conventional video--opinion pair formulation $D=\{V,S\}$ to video--instruction--opinion triplets $D=\{V,I,S\}$, where $V$ denotes the generated video, $I$ represents the textual instruction, and $S$ is the corresponding human evaluation score. As illustrated in Figure~\ref{fig:data_collection}, the standard data collection pipeline typically consists of three stages: (1) instruction collection, (2) video generation using multiple text-to-video models, and (3) human evaluation.

\begin{figure}[!t]
    \centering
    \includegraphics[width=0.75\linewidth]{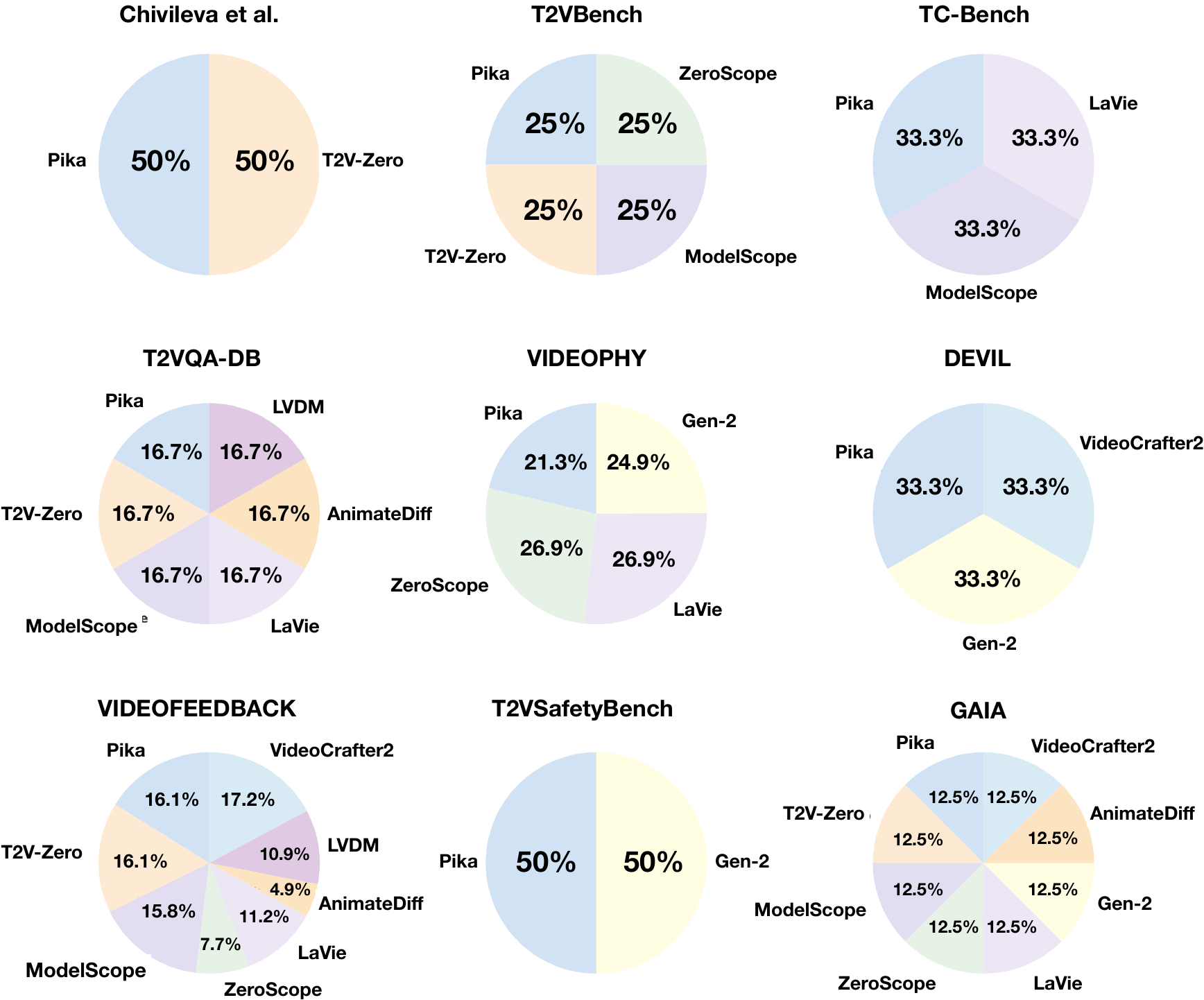}
    \caption{The Proportion of Videos Generated by Each Text-to-video Model.}
    \label{fig:data_statistic}
\end{figure}

Based on their technical evaluation focus, we categorize existing AIGVE benchmarks into five major groups, each serving distinct evaluation purposes. Table~\ref{tab:compre_data} summarizes representative benchmark datasets in terms of scale, annotation protocol, and evaluation focus, while Figure~\ref{fig:data_statistic} presents the distribution of videos generated by different text-to-video models across these benchmarks, highlighting differences in model coverage and dataset composition.

\input{tables/comprehensive_data}

\noindent \textbf{General Large-scale Benchmark Datasets.}
These datasets aim to measure overall visual quality, cross-modal alignment, and temporal coherence of AI-generated videos under diverse real-world prompts. Representative examples include \textbf{T2VQA-DB}~\cite{kou_subjective-aligned_2024-1}, \textbf{AIGC-VQA}~\cite{Lu_2024_CVPR}, and the evaluation framework proposed by \textbf{\citet{chivileva_measuring_2023}}. Such benchmarks are characterized by large-scale prompt collections, outputs from multiple contemporary text-to-video models, and subjective annotations in the form of mean opinion scores or preference ratings.

While their scale enables broad coverage of real-world generation scenarios, the associated annotations are typically coarse-grained, often collapsing complex perceptual and semantic factors into a single or low-resolution score. Moreover, as many of these datasets are collected in the wild with limited control over annotator expertise and evaluation protocols, the resulting labels may exhibit higher variability and noise. Consequently, although these benchmarks are valuable for large-scale training and overall performance comparison, they provide limited diagnostic resolution and are less suitable for isolating fine-grained technical failure modes.
\\\\
\noindent\textbf{Multi-dimensional Benchmarks Datasets.}
To facilitate fine-grained analysis beyond holistic perceptual assessment, several benchmark datasets explicitly decompose AI-generated video quality into multiple semantic, temporal, and stylistic dimensions. \textbf{VBench}~\cite{huang_vbench_2023} exemplifies this category by organizing evaluation into a hierarchical structure spanning 16 dimensions across diverse semantic categories, with dimension-specific prompts and annotations. \textbf{EvalCrafter}~\cite{liu2023evalcrafter},  \textbf{VIDEOFEEDBACK}~\cite{he2024videoscore}, and \textbf{AIGVE-Bench} \cite{xiang2025aigvetoolaigeneratedvideoevaluation} further extend this paradigm by providing multi-aspect annotations that jointly cover visual fidelity, temporal coherence, motion dynamics, cross-modal alignment, and content or factual consistency across large collections of videos generated by contemporary text-to-video models.

More recently, benchmark datasets have begun to enrich multi-aspect numerical annotations with \emph{aspect-wise natural language comments}. \textbf{AIGVE-BENCH~2} \cite{liu2025aigvemacsunifiedmultiaspectcommenting} is a large-scale human-annotated dataset comprising 2,500 AI-generated videos paired with both numerical scores and detailed explanatory comments across nine evaluation aspects, resulting in a total of 22,500 aspect-level comments that explicitly justify the assigned scores. Similarly, \textbf{VIDEOFEEDBACK2} \cite{he2025videoscore2thinkscoregenerative} extends earlier multi-dimensional datasets by annotating each video with scores and accompanying rationale text along three core dimensions: visual quality, text-to-video alignment, and physical or commonsense consistency, yielding over 27,000 annotated videos with score–comment pairs.

By explicitly structuring annotations at the aspect level and augmenting scalar scores with corresponding comments, these datasets provide richer supervision signals than score-only benchmarks, capturing not only \textit{how well} a video performs along each dimension but also \textit{why it receives a given evaluation}.
\\\\
\noindent\textbf{Temporal and Motion-centric Benchmark Datasets.}
Temporal modeling remains a core challenge for text-to-video generation, motivating benchmarks that explicitly emphasize motion fidelity, event progression, and temporal consistency. \textbf{T2VBench}~\cite{Ji_2024_CVPR} and \textbf{DEVIL}~\cite{liao2024evaluation} focus on motion dynamics and controllability through prompts enriched with temporal attributes and motion-related lexicons. \textbf{TC-Bench}~\cite{feng2024tc} further targets temporal compositionality by evaluating attribute transitions, object relation dynamics, and environment changes across time using paired prompts and reference videos.
\textbf{GAIA}~\cite{chen2024gaiarethinkingactionquality} concentrates on action-centric temporal evolution. The dataset comprises large-scale human annotations assessing action completeness, subject consistency, and action–scene interaction, emphasizing whether actions unfold coherently and plausibly over time rather than static visual fidelity.

Collectively, these benchmarks provide structured datasets for evaluating long-range temporal consistency and dynamic realism in generated videos.
\\\\
\noindent\textbf{Real-World--Oriented Evaluation Benchmarks.}
Beyond perceptual quality and temporal coherence, a growing set of evaluation benchmarks focuses on assessing how well text-to-video models reflect real-world correctness and real-world impact. Rather than measuring visual realism alone, these datasets probe whether generated videos adhere to physical laws, respect safety constraints, and accurately represent real-world entities, environments, and knowledge.

\textbf{VIDEOPHY}~\cite{bansal2024videophy} targets physical commonsense in video generation by constructing prompts involving multi-state matter interactions, such as solid--fluid and fluid--fluid dynamics, and annotating generated videos for physical plausibility and semantic fidelity. \textbf{T2VSafetyBench}~\cite{miao2024t2vsafety} evaluates safety-related behaviors using adversarial prompts and corresponding video generations labeled across multiple risk dimensions, including explicit content, violence, discrimination, and temporal safety concerns, highlighting potential real-world deployment risks.

\textbf{GEOATTRACTION-500}~\cite{liu2026videogenerationmodelsgeographically} further extends real-world–oriented evaluation to geographically grounded visual knowledge. The dataset comprises 500 globally distributed tourist attractions paired with curated reference images, structured textual prompts, and attraction-level metadata such as geographic region and popularity. By focusing on whether generated videos faithfully capture location-specific visual identity and structural characteristics, GEOATTRACTION-500 enables systematic analysis of real-world knowledge coverage and geographic equity in text-to-video models.
\textbf{ADGV-Bench}~\cite{ADGVE} introduces a driving-specific benchmark dataset comprising AI-generated driving videos annotated with human judgments across perceptual fidelity, temporal coherence, and physical plausibility dimensions, enabling systematic evaluation of video generation models in the autonomous driving domain.
Collectively, these benchmarks complement quality-centric evaluations by assessing aspects of generation that are directly tied to real-world correctness, societal impact, and deployment considerations, which are not adequately captured by perceptual quality metrics alone.

In summary, existing AIGVE benchmarks are designed to serve distinct evaluation objectives and should be selected accordingly. General-purpose large-scale datasets support holistic quality assessment, hierarchical and multi-dimensional benchmarks enable fine-grained diagnostic analysis, temporal-centric benchmarks focus on motion and long-range consistency, and real-world–oriented benchmarks evaluate physical plausibility, safety, and grounded real-world knowledge. As no single benchmark captures all these aspects, comprehensive evaluation of AI-generated videos typically requires combining multiple benchmark categories.

\begin{figure}[t]
\centering
\footnotesize
\scalebox{0.90}{

\begin{forest}
for tree={
    grow'=0,
    parent anchor=east,
    child anchor=west,
    anchor=west,
    draw,
    line width=0.8pt,
    edge={gray!70, line width=0.8pt},
    rounded corners=8pt,
    inner xsep=8pt,
    inner ysep=4pt,
    minimum height=8mm,
    align=left,
    l sep=12mm,
    s sep=7mm,
    edge path={
        \noexpand\path[\forestoption{edge}]
        (!u.east) -- +(6pt,0) |- (.west)\forestoption{edge label};
    },
},
root/.style={
    draw=black!70,
    fill=gray!15,
},
metric/.style={
    draw=red!70,
    fill=red!10,
},
model/.style={
    draw=green!60!black,
    fill=green!10,
},
metriccite/.style={
    draw=red!70,
    fill=red!5,
},
modelcite/.style={
    draw=green!60!black,
    fill=green!5,
},
where level=0{root}{},
[AIGVE, root
    [Metric Collection Evaluation, metric
        [{{\cite{huang_vbench_2023}, \cite{liu2023evalcrafter}}}, metriccite]
    ]
    [Modeling-Based Evaluation, model
        [Holistic Evaluation, model
            [{{\cite{kou_subjective-aligned_2024-1}, \cite{Lu_2024_CVPR}}}, modelcite]
        ]
        [Multi-aspect Evaluation, model
            [{{\cite{he2024videoscore}, \cite{he2025videoscore2thinkscoregenerative}, \cite{liu2025aigvemacsunifiedmultiaspectcommenting}}}, modelcite]
        ]
        [Temporal Evaluation, model
            [{{\cite{liao2024evaluation}, \cite{feng2024tc}, \cite{FVMD}}}, modelcite]
        ]
        [Real-world Capability, model
            [{{\cite{miao2024t2vsafety}, \cite{liu2026videogenerationmodelsgeographically},\cite{ADGVE}}}, modelcite]
        ]
    ]
]
\end{forest}

}

\caption{Taxonomy of Current AIGVE Methods.}
\label{fig:aigve_tax}
\end{figure}

\subsection{Evaluation Methods for AI-Generated Video Evaluation}
\label{sec_comprehensive}

With the availability of diverse benchmark datasets that align with various evaluation aspects, research in AIGVE has seen rapid growth. This section categorizes and summarizes the advancements in AIGVE research. Figure \ref{fig:aigve_tax} summarizes the existing methods that are designed for AIGVE.
\subsubsection{Metric Collection Evaluation}
In the early stages of AI-generated video research, as the demand for robust evaluation methods grew, researchers recognized the complexity of providing a comprehensive assessment. Instead of assigning a single score to summarize the quality of the generated video, they began to decompose the evaluation process into multiple aspects. Each aspect was assessed using established metrics, allowing for a more granular and accurate evaluation of AI-generated videos.


\textbf{VBench} \cite{huang_vbench_2023} presents a comprehensive evaluation framework that decomposes video generation quality into 16 hierarchical dimensions across Video Quality and Video Condition Consistency axes. The Video Quality metrics incorporate Subject and Background Consistency (via DINO \cite{caron2021emergingpropertiesselfsupervisedvision} and CLIP \cite{radford2021learning}), temporal characteristics (flickering and motion smoothness), Dynamic Degree (RAFT \cite{teed2020raftrecurrentallpairsfield}), and perceptual attributes (LAION \cite{schuhmann2022laion}, MUSIQ \cite{ke2021musiqmultiscaleimagequality}). The Condition Consistency axis evaluates prompt adherence through object detection (GRiT \cite{wu2022grit}), action conformity (UMT \cite{li2023unmasked}), compositional elements, scene concordance (Tag2Text \cite{huang2023tag2text}), and stylistic coherence (CLIP \cite{radford2021learning}, ViCLIP \cite{wang2023internvid}), demonstrating strong correlation with human perceptual judgments while providing granular insights into model capabilities. Similarly, \textbf{EvalCrafter} \cite{liu2023evalcrafter} presents a comprehensive evaluation framework for text-to-video models structured across four fundamental dimensions: visual quality, text-video alignment, motion quality, and temporal consistency. The framework employs multiple complementary metrics: visual quality assessment through aesthetic and technical ratings alongside inception score \cite{salimans2016improvedtechniquestraininggans} for content diversity; text-video alignment via CLIP-Score \cite{hessel2022clipscore}, BLIP-BLEU \cite{li2023blip, Papineni2002BleuAM}, and a novel SD-Score \cite{rombach_high-resolution_2022} benchmarked against stable diffusion outputs; motion quality through Action-Score and Flow-Score metrics; and temporal consistency evaluated via warping error and cross-frame semantic coherence measures. This multi-faceted approach enables systematic assessment of generative video models across perceptual and technical quality dimensions.

Both works establish a foundational framework for evaluating AI-generated video models, demonstrating strong correlations with human judgment. However, the need for multiple individual metrics in these frameworks presents challenges in integrating them into a unified pipeline. Additionally, evaluating a single video across a sequence of metrics is time-consuming. Thus, there remains a need for a streamlined, unified evaluation method.

\subsubsection{Modeling-Based Evaluation Methods}

Building upon large-scale annotated benchmarks, recent research on AI-generated video evaluation has increasingly adopted modeling-based approaches that learn to approximate human judgment directly from data. Rather than relying on fixed metrics, these methods differ in how evaluation signals are modeled and structured, ranging from holistic quality prediction to multi-aspect, temporal, and real-world–oriented capability assessment. This progression reflects an increasing emphasis on interpretability, temporal reasoning, and real-world relevance in learned video evaluators.
\\\\
\noindent\textbf{Holistic Evaluation Modeling.}
Holistic evaluators aim to produce an overall assessment of a generated video by jointly considering perceptual fidelity and text--video alignment. \textbf{T2VQA}~\cite{kou_subjective-aligned_2024-1} learns an end-to-end quality predictor trained on T2VQA-DB, integrating signals from both alignment-oriented visual–language representations and fidelity-oriented video features to regress an overall quality score. A\textbf{IGC-VQA}~\cite{Lu_2024_CVPR} similarly targets holistic assessment, combining multiple complementary perceptual cues.
\\\\
\noindent\textbf{Multi-aspect and Comment-aware Evaluation Modeling.}
To better reflect how humans evaluate videos along distinct criteria, multi-aspect evaluators explicitly predict structured feedback across multiple dimensions rather than collapsing judgments into a single score. \textbf{VideoScore}~\cite{he2024videoscore}, trained on the VIDEOFEEDBACK dataset, represents a step in this direction by modeling fine-grained human feedback across multiple evaluation aspects. \textbf{VideoScore2} \cite{he2025videoscore2thinkscoregenerative} further extends this line by introducing multi-aspect evaluation with explicit reasoning over key dimensions such as visual quality, text alignment, and physical consistency. In contrast to earlier score-only models, VideoScore2 produces aspect-wise scores and accompanying natural language comments, and is trained with reinforcement learning to better align its outputs with structured human feedback. \textbf{AIGVE-MACS}~\cite{liu2025aigvemacsunifiedmultiaspectcommenting} similarly advances interpretability by jointly predicting aspect-wise scores and corresponding natural language comments, leveraging large-scale human annotations that pair ratings with textual justifications. By producing structured, aspect-level outputs together with explanatory comments, these models provide more interpretable and diagnostically useful evaluations than holistic predictors.
\\\\
\noindent\textbf{Temporal evaluation modeling.}
Complementary to holistic and multi-aspect quality assessment, a class of modeling-based evaluators focuses specifically on temporal capabilities that are central to video generation but insufficiently captured by static perceptual metrics. \textbf{DEVIL}~\cite{liao2024evaluation} exemplifies this direction by modeling motion dynamics through the assessment of dynamics range, controllability, and dynamics-based quality, using prompts with graded motion complexity to probe a model’s ability to generate and control temporal variation. \textbf{TC-Bench}~\cite{feng2024tc} further targets temporal compositionality by evaluating whether attribute transitions, object relations, and environmental changes described in text prompts are correctly realized and completed over time. These models align closely with temporal-centric benchmark datasets and provide structured evaluation of motion realism and long-range temporal coherence. \textbf{FVMD}~\cite{FVMD} instead models temporal quality through explicit motion features, capturing velocity and acceleration statistics from tracked trajectories and comparing their distributions via a Fréchet distance. By focusing on motion consistency and physically plausible dynamics, it better reflects temporal coherence and aligns more closely with human judgments than prior metrics. These models align closely with temporal-centric benchmark datasets and provide structured evaluation of motion realism and long-range temporal coherence.
\\\\
\noindent\textbf{Real-world--oriented Capability Evaluation Modeling.}
In parallel, another category of modeling-based evaluators emphasizes real-world grounding and deployment-relevant constraints that extend beyond visual quality. \textbf{T2VSafetyBench}~\cite{miao2024t2vsafety} adopts a modeling-based framework to assess safety-related risks across multiple content dimensions, including explicit, violent, discriminatory, and misleading content, reflecting concerns critical to real-world usage. \textbf{GAP} (Geo-Attraction Landmark Probing) \cite{liu2026videogenerationmodelsgeographically} focuses on real-world grounding by modeling whether generated videos faithfully capture geographically grounded visual knowledge, comparing generated content against curated reference imagery of real-world landmarks. 
\textbf{ADGVE}~\cite{ADGVE} trains a domain-specific evaluator that jointly assesses perceptual fidelity, temporal coherence, and physical plausibility of AI-generated driving videos for downstream autonomous driving perception tasks
such as multimodal 3D object detection\cite{FusionViT, 3DifFusionDet, OVODA} and efficient scene understanding \cite{EffiPerception}.
By explicitly evaluating adherence to physical, social, and geographic constraints, these models complement perceptual and temporal evaluators and enable assessment of real-world impact and reliability.

Overall, modeling-based AIGVE methods have evolved from producing coarse, holistic quality scores toward structured, aspect-aware, and capability-driven evaluation. Holistic models remain effective for high-level system comparison, while multi-aspect and comment-aware approaches provide improved interpretability, and temporal and real-world–oriented models enable targeted assessment of dynamic behavior and deployment-relevant constraints. These paradigms are complementary rather than mutually exclusive, and comprehensive evaluation of AI-generated videos increasingly requires combining modeling approaches across these categories.

\subsection{Foundations of AI-Generated Video Evaluation}
Although still emerging, AIGVE builds on two key foundations: alignment with human perception and alignment with human instructions. A solid understanding of these two foundational aspects is crucial for comprehensively developing AIGVE. 

In the subsequent sections of this survey, we introduce the research works related to these two important foundation aspects and give further prospects based on the ground knowledge of these two aspects.

Section \ref{sec_preception} discusses alignment with human perception. This section investigates a detailed survey on benchmark datasets as well as existing quality evaluation methods, offering a clear picture of how the field of evaluating video perception quality evolved over the last decade.

Section \ref{sec_instruction} covers alignment with human instructions. This section summarizes the prior representative work in terms of the alignment between video content and human instructions to introduce the previous development trajectory of this field and how it has been influenced by the emergence of large language models.

Finally, Section \ref{sec_future} summarizes future perspectives in the field, identifying key challenges and potential research opportunities. We discuss the integration of vision language models for video evaluation, improvements in score interpretability, and the embedding of ethical and safety considerations within AIGVE frameworks.

%% file: tables/comprehensive_data.tex
\begin{table}[!t]
\centering
\fontsize{8pt}{11pt}\selectfont
\caption{Summary of recent AI-Generated Video Evaluation Benchmark datasets: '\# Video' is the number of unique videos, '\# Frames' the frames per video, '\# Prompt' the unique prompts, '\# Category' the number of video categories, '\# Generators' the number of video generation models, '\# Evaluators' the number of human evaluators, and 'Aspect' the evaluation focus.}

\scalebox{0.73}{
\begin{tabular}{l l l l l l l l l l c}
\toprule
\textbf{Dataset} & \textbf{Year} & \textbf{\# Video} & \textbf{\# Frames} & \textbf{Resolution} & \textbf{\# Prompt} & \textbf{\# Category} & \textbf{\# Generators} & \textbf{\# Evaluators} & \textbf{Aspect} & \textbf{Link} \\
\midrule
EvalCrafter \cite{liu2023evalcrafter} & 2023 & 2,500 & 8 & > 512p & 500 & 4 & 5 & 3 & general & \href{https://evalcrafter.github.io}{\includegraphics[width=0.03\textwidth]{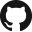}} \\
VBench \cite{huang_vbench_2023} & 2023 & - & > 6 & > 240p & 1746 & 24 & 8 & - & general & \href{https://vchitect.github.io/VBench-project/}{\includegraphics[width=0.03\textwidth]{figures/github_logo.png}} \\
\citet{chivileva_measuring_2023} & 2023 & 1,005 & > 8 & > 128p & 201 & 2 & 5 & 24 & naturalness & \includegraphics[width=0.03\textwidth]{figures/github_logo.png} \\
T2VQA-DB \cite{kou_subjective-aligned_2024-1} & 2024 & 10,000 & 16 & 512p & 1,000 & 7 & 9 & 27 & general & \href{https://github.com/QMME/T2VQA}{\includegraphics[width=0.03\textwidth]{figures/github_logo.png}} \\
T2VBench \cite{Ji_2024_CVPR} & 2024 & 5,000 & > 16 & > 256p & 1,600 & 16 & 3 & 3/data & temporal & \href{https://ji-pengliang.github.io/T2VBench}{\includegraphics[width=0.03\textwidth]{figures/github_logo.png}} \\
VIDEOPHY \cite{bansal2024videophy} & 2024 & 9,300 & > 25 & > 240p & 688 & 3 & 9 & - & physical & \href{https://videophy.github.io}{\includegraphics[width=0.03\textwidth]{figures/github_logo.png}} \\
TC-Bench \cite{feng2024tc} & 2024 & 817 & 16 & > 256p & 150 & 3 & 5 & 8 & composition & \href{https://weixi-feng.github.io/tc-bench/}{\includegraphics[width=0.03\textwidth]{figures/github_logo.png}} \\
T2VSafetyBench \cite{miao2024t2vsafety} & 2024 & 17,600 & > 25 & > 240p & 4,400 & 12 & 4 & 60 & safety & \includegraphics[width=0.03\textwidth]{figures/github_logo.png} \\
VIDEOFEEDBACK \cite{he2024videoscore} & 2024 & 37,600 & > 8 & > 256p & 44,500 & 5 & 11 & 20 & general & \href{https://tiger-ai-lab.github.io/VideoScore/}{\includegraphics[width=0.03\textwidth]{figures/github_logo.png}} \\
AIGC-VQA \cite{Lu_2024_CVPR} & 2024 & 10,000 & 16 & 512p & 1,000 & 3 & 10 & 20 & general & \includegraphics[width=0.03\textwidth]{figures/github_logo.png} \\
DEVIL \cite{liao2024evaluation} & 2024 & 800 & > 16 & > 240p & 4,800 & 5 & 6 & 6 & dynamics & \href{https://github.com/MingXiangL/DEVIL}{\includegraphics[width=0.03\textwidth]{figures/github_logo.png}} \\
GAIA \cite{chen2024gaiarethinkingactionquality} & 2024 & 9,180 & > 4 & > 256p & 510 & 3 & 18 & 54 & action & \href{https://github.com/zijianchen98/GAIA}{\includegraphics[width=0.03\textwidth]{figures/github_logo.png}} \\

AIGVE-Bench \cite{xiang2025aigvetoolaigeneratedvideoevaluation} & 2025 & 2500 & > 4 & > 480p & 500 & 2 & 5 & - & general & \href{https://huggingface.co/datasets/xiaoliux/AIGVE-Bench}{\includegraphics[width=0.03\textwidth]{figures/github_logo.png}} \\
\bottomrule
\end{tabular}
}
\vspace{-2mm}

\label{tab:compre_data}
\end{table}

%% file: content/4-vqa.tex
\section{Alignment with Human Perception}
\label{sec_preception}

A key objective of AI-generated video is to produce content that closely aligns with human perception. Over the past decade, numerous benchmark datasets and evaluation techniques have been developed to assess AI-generated videos based on criteria related to human perception. In this section, we introduce several representative benchmark datasets, evaluation methods, and associated metrics.

\subsection{Benchmark Datasets}


A number of datasets have been curated for the development and validation of video quality assessment models. This section details various collections of video sequences with different characteristics and their corresponding quality scores. These videos vary in resolution, duration, distortions, and the environments in which they were assessed, as well as in the methodologies used for assessing video quality, including crowdsourcing platforms, controlled lab environments, and various subjective quality metrics. The evaluation metrics employed across these datasets include Mean Opinion Score (MOS) \cite{streijl2016MOS}, Differential Mean Opinion Score (DMOS) \cite{sheikh2006statistical}, Peak Signal-to-Noise Ratio (PSNR) \cite{shannon1948mathematical}, Structural Similarity Index (SSIM) \cite{wang2004image}, and no-reference metrics such as BRISQUE \cite{mittal2012no} and NIQE \cite{mittal2013making}, providing a comprehensive assessment of video quality from multiple perspectives. A summary of the benchmark datasets discussed in this section is presented in Table \ref{tab:percep_dataset}.

To reduce fragmented descriptions and provide a clearer map of the landscape, we group benchmark datasets by their data origin and collection protocol, shown in Table~\ref{tab:percep_dataset_tax}.
This taxonomy highlights what each dataset family is best suited to evaluate (and what it may miss), which is helpful when interpreting results across metrics and benchmarks.

\begin{table}[!t]
\centering
\caption{Summary of Databases for Video Quality Assessment. 'Cont.' denotes the number of unique video contents. '\#Total' represents the total number of videos. 'Dur.' indicates the video duration in seconds. '\#Subj.' refers to the number of subjects involved. 'Env.' specifies the environment for the subjects. 'Crowd' stands for crowdsourcing.}
\scalebox{0.78}{
\begin{tabular}{l l l l l l l l l l l}
\toprule
\textbf{Database} & \textbf{Year} & \textbf{\#Cont.} & \textbf{\#Total} & \textbf{Resolution} & \textbf{Dur.} & \textbf{Distortions} & \textbf{\#Subj.} & \textbf{Env.} & \textbf{Link} \\
\midrule

CVD2014 \cite{nuutinen2016cvd2014} & 2014 & 5 & 234 & 480p, 720p & 10-25 & In-capture & 210 & In-Lab & \href{https://qualinet.github.io/databases/video/cvd2014_video_database/}{\includegraphics[width=0.03\textwidth]{figures/github_logo.png}} \\
KoNViD-1k \cite{7965673} & 2017 & 1200 & 1200 & 540p & 8 & In-the-wild & 642 & Crowd & \href{https://database.mmsp-kn.de/konvid-1k-database.html}{\includegraphics[width=0.03\textwidth]{figures/github_logo.png}} \\
LIVE-VQC \cite{8463581} & 2018 & 585 & 585 & 1080p, 240p & 10 & In-the-wild & 4776 & Crowd & \href{https://live.ece.utexas.edu/research/LIVEVQC/index.html}{\includegraphics[width=0.03\textwidth]{figures/github_logo.png}} \\
YouTube-UGC \cite{wang2019youtube} & 2019 & 1500 & 1500 & 4K, 360p & 20 & In-the-wild & >8000 & Crowd & \href{https://media.withyoutube.com}{\includegraphics[width=0.03\textwidth]{figures/github_logo.png}}\\
SPSS \cite{9318149} & 2020 & 14 & 224 & 1080p & N/A & In-the-wild & 19 & Crowd & \href{https://data.mendeley.com/datasets/f9xd5zp8mp/1}{\includegraphics[width=0.03\textwidth]{figures/github_logo.png}} \\
UGC-VIDEO \cite{li2020ugcvideo} & 2020 & 50 & 550 & 720p & 10 & UGC+compression & 30 & In-lab & \href{}{\includegraphics[width=0.03\textwidth]{figures/github_logo.png}} \\
ETRI-LIVE-STSVQ \cite{Lee2022} & 2021 & 15 & 437 & 4K & 5-7 & In-the-wild & 34 & In-lab & \href{https://live.ece.utexas.edu/research/ETRI-LIVE_STSVQ/index.html}{\includegraphics[width=0.03\textwidth]{figures/github_logo.png}}\\
LIVE-APV \cite{Shang2022} & 2021 & 33 & 315 & 1080p, 4K & 7 & In-the-wild & 40 & In-lab & \href{https://live.ece.utexas.edu/research/LIVE_APV_Study/apv_index.html}{\includegraphics[width=0.03\textwidth]{figures/github_logo.png}}\\
LIVE-YT-HFR \cite{madhusudana2021subjective} & 2021 & 16 & 480 & 4K & 10 & In-the-wild & 85 & In-lab & \href{https://live.ece.utexas.edu/research/LIVE_YT_HFR/LIVE_YT_HFR/index.html}{\includegraphics[width=0.03\textwidth]{figures/github_logo.png}}\\
LIVE-LSVQ \cite{Patch_VQ} & 2022 & 39075 & 39075 & 1080p & 5-12 & In-the-wild & 6284 & Crowd & \href{https://github.com/baidut/PatchVQ}{\includegraphics[width=0.03\textwidth]{figures/github_logo.png}}\\
MSU CVQAD \cite{NEURIPS2022_59ac9f01} & 2022 & 2500 & 2486 & 360p-1080p & 10, 15 & Compression & 10800 & Crowd & \href{https://videoprocessing.ai/datasets/cvqad.html}{\includegraphics[width=0.03\textwidth]{figures/github_logo.png}}\\
M-VCM \cite{10095711} & 2023 & 10 & 1628 & 1080p & 6 & In-capture & N/A & Crowd & \href{https://github.com/microsoft/Video_Call_MOS?tab=readme-ov-file}{\includegraphics[width=0.03\textwidth]{figures/github_logo.png}}\\

\bottomrule

\end{tabular}

}
\vspace{-3mm}
\label{tab:percep_dataset}
\end{table}

\begin{table}[!t]
\centering
\caption{A categorical view of benchmark datasets for alignment with human perception.}
\label{tab:percep_dataset_tax}
\scalebox{0.9}{%
\begin{tabular}{l p{4.5cm} p{6.4cm}}
\toprule
\textbf{Category} & \textbf{Representative datasets} & \textbf{Typical characteristics} \\
\midrule
UGC datasets & KoNViD-1k, LIVE-VQC, YouTube-UGC, LIVE-LSVQ & Authentic distortions; broad content diversity \\
Controlled datasets & CVD2014, UGC-VIDEO, ETRI-LIVE-STSVQ, LIVE-APV, LIVE-YT-HFR & Controlled viewing; structured distortion controls \\
Application-specific datasets & SPSS, MSU CVQAD, M-VCM & Surveillance/compression/network degradations \\
\bottomrule
\end{tabular}}
\vspace{-2mm}
\end{table}

\subsubsection{UGC datasets}


UGC datasets are collected from real-world user videos and are typically labeled via crowdsourced subjective studies, thus containing authentic, mixed distortions with high content diversity.
They offer strong ecological validity, but their distortions are often uncontrolled and entangled, which can make diagnosis difficult.

Representative benchmarks include \textbf{KoNViD-1k} \cite{7965673}, which curates user-generated content from YFCC100m and provides mean opinion scores for large-scale no-reference quality modeling. Its diversity and authentic distortion mixtures make it a common starting point for evaluating generalization beyond laboratory distortions. \textbf{LIVE-VQC} \cite{8463581} further targets consumer-captured videos and emphasizes realistic capture conditions, including frequent camera motion and exposure variation that challenge both spatial and temporal quality predictors. It is often used to stress-test robustness to the kinds of artifacts that are difficult to synthesize cleanly. \textbf{YouTube-UGC} \cite{wang2019youtube} expands coverage with broader content types and resolution ranges from online videos, and it is frequently adopted to benchmark cross-content stability of learned perceptual representations. Compared with smaller UGC sets, it supports more reliable comparison among methods that rely on large-scale training. \textbf{LIVE-LSVQ} \cite{Patch_VQ} scales UGC evaluation to a substantially larger collection and additionally supports global--local quality analysis through patch-level supervision. This design makes it particularly useful when studying how local defects and global impressions interact under mixed degradations.

\subsubsection{Controlled datasets}

Controlled datasets are created under laboratory settings or systematic capture conditions, where distortion factors (e.g., compression level, frame rate, resolution, capture artifacts) are explicitly manipulated.
They enable cleaner factor-wise analysis and fair comparisons, but may cover a narrower content distribution and can be less representative of complex in-the-wild degradations. 

\textbf{CVD2014}~\cite{nuutinen2016cvd2014} is a dataset that focuses on capture-related distortions under controlled viewing and offers a structured setting for isolating specific degradation sources. It is frequently used when the goal is to evaluate sensitivity to camera and acquisition artifacts rather than post-processing alone. \textbf{UGC-VIDEO} \cite{li2020ugcvideo} constructs controlled transcoding and compression variations from short-video sources, enabling direct analysis of how codec and bitrate changes affect perceived quality. It provides a bridge between authentic content and controlled distortion manipulation, which is useful for studying robustness under platform-style processing. \textbf{ETRI-LIVE-STSVQ} \cite{Lee2022} systematically varies spatial and temporal sampling together with compression, making it well suited for analyzing trade-offs between motion fidelity and spatial detail. Its design supports targeted evaluation of temporal degradation patterns that may be underrepresented in many purely spatial-focused benchmarks. \textbf{LIVE-APV} \cite{Shang2022} emphasizes perceptual artifacts such as judder, flicker, and frame drops, offering a controlled resource to test whether metrics respond to temporal disruptions that strongly affect human viewing experience. It is especially relevant when assessing whether a method can distinguish temporally unstable outputs from visually sharp but motion-inconsistent ones. \textbf{LIVE-YT-HFR} \cite{madhusudana2021subjective} concentrates on high-frame-rate content with systematic frame-rate and compression settings, and it is commonly used to evaluate sensitivity to frame-rate changes and high-motion scenarios. Together, these controlled datasets complement UGC benchmarks by enabling clearer attribution from performance differences to specific distortion factors.

\subsubsection{Application-specific datasets}

Application-specific datasets focus on degradations arising from practical pipelines or domain constraints (e.g., surveillance, streaming, network impairments, device-specific artifacts), reflecting real deployment scenarios.
They are valuable for measuring robustness under realistic system effects, but their scope can be domain-dependent and may not generalize to broader AIGV settings.

Typical datasets consist of \textbf{SPSS} \cite{9318149}, which targets surveillance scenarios and emphasizes distortions that arise in security and monitoring pipelines. It is valuable for testing whether perceptual metrics remain reliable when video content is constrained by domain-specific viewpoints, lighting, and device characteristics. \textbf{MSU CVQAD} \cite{NEURIPS2022_59ac9f01} evaluates large-scale codec and transmission variations and is notable for using pairwise comparisons that are later aggregated into quality scores. This setting is well aligned with practical decision-making, where systems often need to choose between two candidate streams under bandwidth constraints. \textbf{M-VCM} \cite{10095711} focuses on video conferencing quality under diverse network environments and captures impairments that occur in real communication systems. Its subjective study design follows standardized procedures, and it connects naturally to engineering practice through the ITU-T P.910 toolkit \cite{naderi2024}. Compared with general UGC benchmarks, these application-specific datasets emphasize system-induced degradations and help validate whether evaluation methods are suitable for deployment-facing scenarios.

\begin{figure*}[t]
\centering
\scriptsize
\scalebox{0.82}{
\begin{forest}
for tree={
    grow'=0,
    anchor=west,
    parent anchor=east,
    child anchor=west,
    edge={gray!70, line width=0.9pt},
    rounded corners=6pt,
    l sep=10mm,
    s sep=4mm,
    inner xsep=6pt,
    inner ysep=4pt,
    minimum height=7mm,
    edge path={
        \noexpand\path[\forestoption{edge}]
        (!u.east) .. controls +(8pt,0) and +(-8pt,0) .. (.west)\forestoption{edge label};
    },
    font=\normalsize,
    draw=none,
},
root/.style={
    draw=black!70,
    fill=gray!15,
    rounded corners=8pt,
    font=\bfseries,
},
stat/.style={
    draw=red!70,
    fill=red!10,
    font=\bfseries,
},
statcite/.style={
    draw=red!70,
    fill=red!5,
},
cnn/.style={
    draw=green!60!black,
    fill=green!10,
    font=\bfseries,
},
cnncite/.style={
    draw=green!60!black,
    fill=green!5,
},
trans/.style={
    draw=blue!65!black,
    fill=blue!10,
    font=\bfseries,
},
transcite/.style={
    draw=blue!65!black,
    fill=blue!5,
},
clip/.style={
    draw=orange!85!black,
    fill=orange!12,
    font=\bfseries,
},
clipcite/.style={
    draw=orange!85!black,
    fill=orange!5,
},
llm/.style={
    draw=purple!70!black,
    fill=purple!10,
    font=\bfseries,
},
llmcite/.style={
    draw=purple!70!black,
    fill=purple!5,
},
mid/.style={
    font=\normalsize,
},
leaf/.style={
    font=\normalsize,
},
[Alignment with Human Perception, root
    [Statistical-based VQA, stat
        [{{\cite{VMAF}, \cite{EPooling}}}, statcite, leaf]
    ]
    [CNN-based VQA, cnn
        [Temporal modeling methods, cnn
            [{{\cite{VSFA}, \cite{chung2014empiricalevaluationgatedrecurrent}, \cite{MDTVSFA}, \cite{GSTVQA}}}, cnncite, leaf]
        ]
        [Feature fusion methods, cnn
            [{{\cite{VIDEVAL}, \cite{RAPIQUE}, \cite{FR_NR_VQA}}}, cnncite, leaf]
        ]
        [Multi-scale analysis methods, cnn
            [{{\cite{Patch_VQ}, \cite{Pro.}, \cite{HVS_5M}, \cite{MultiLevelFusion}, \cite{NTIRE2024_SUGC}}}, cnncite, leaf]
        ]
        [Quality disentanglement methods, cnn
            [{{\cite{wu_exploring_2023-1}, \cite{MD_VQA}, \cite{Light_VQA}, \cite{Ada_DQA}, \cite{BVQA}}}, cnncite, leaf]
        ]
    ]
    [Transformer-based VQA, trans
        [Attention-based methods, trans
            [{{\cite{StarVQA}, \cite{FANet}, \cite{VQT}, \cite{SSL_VQA}, \cite{MVQA_Mamba}}}, transcite, leaf]
        ]
        [Hierarchical methods, trans
            [{{\cite{Zoom_VQA}, \cite{SB_VQA}, \cite{COVER}, \cite{UnifiedVQA}}}, transcite, leaf]
        ]
        [Prior-guided methods, trans
            [{{\cite{Light_VQA_plus}, \cite{PTM_VQA}, \cite{Priorformer}, \cite{ReLaX_VQA}}}, transcite, leaf]
        ]
    ]
    [CLIP-based VQA, clip
        [Semantic-driven methods, clip
            [{{\cite{BVQI}, \cite{MaxVQA}}}, clipcite, leaf]
        ]
        [UGC-focused methods, clip
            [{{\cite{KSVQE}, \cite{RQ_VQA}}}, clipcite, leaf]
        ]
        [AIGC-oriented methods, clip
            [{{\cite{UGVQ}, \cite{NTIRE2024_AIGCQA}, \cite{CLIPVQA}}}, clipcite, leaf]
        ]
    ]
    [LLM-based VQA, llm
        [Instruction-tuned methods, llm
            [{{\cite{QAlign}, \cite{LMM_VQA}, \cite{VQA2}}}, llmcite, leaf]
        ]
        [Efficient LMM-based methods, llm
            [{{\cite{QCLIP_2025}}}, llmcite, leaf]
        ]
    ]
]
\end{forest}

}

\caption{Taxonomy of Alignment with Human Perception Methods.}
\label{fig:human_perception_vqa_forest}
\end{figure*}

\subsection{Evaluation Methods}

Video Quality Assessment (VQA) algorithms aim to predict video quality in alignment with human perception. 
The rapid expansion of social media platforms, such as YouTube, Facebook, and TikTok, has driven a surge in no-reference user-generated video content. 
While professional-generated content has received less attention, possibly due to copyright concerns, much of the recent research in VQA has focused on no-reference user-generated content. 
For no-reference (NR) VQA, a basic approach involves assessing the quality of each frame using NR-IQA methods and aggregating the results to produce an overall video quality score. 
However, compared to NR-IQA, NR-VQA must account for temporal distortions, which adds complexity by requiring an understanding of time-dependent quality degradations. 

Early NR-VQA algorithms are often designed to address specific types of distortions, such as those caused by transmission or compression artifacts \cite{Codec_Analysis, Channel_Induced_Distortion, coding_artifacts, Predictive, Degraded_and_Enhanced, high_definition}. 
    Their methods are usually statistical-based, which leverage handcrafted machine learning matrices, such as SVM\cite{SVM}, to train regression models for perceptual quality prediction\cite{VMAF, Codec_Analysis, Blind_Prediction, A_Completely_Blind, Spatiotemporal_Statistics, Spatio_Temporal_Measures}.
More contemporary methods adopt complex neural networks such as CNN\cite{CNN}, ViT\cite{ViT}, or Swin-Transformer\cite{swim} to extract a vast array of perceptually relevant features\cite{MDTVSFA, FR_NR_VQA,GSTVQA, FANet, Zoom_VQA, VQT}, processing videos in an end-to-end manner. 
    Some of these methods extract multi-scale features to capture both global and local information, which helps in modeling different levels of perceptual quality\cite{HVS_5M, Light_VQA, Zoom_VQA}. 
Given the subjective nature of video quality assessment, recent models are increasingly looking at ways to decompose a single no-reference VQA score into multiple dimensions \cite{COVER, Pro., wu_exploring_2023-1, MD_VQA}, such as aesthetics feature, semantic feature, distortion feature, motion feature, etc. 
    They may adopt domain-fusion or knowledge transfer to incorporate information from different feature dimensions, enhancing the overall understanding of video quality \cite{Ada_DQA, SSL_VQA, PTM_VQA}. We summarize the VQA methods discussed in this section chronologically in Figure \ref{fig:human_perception_vqa_forest}.

To provide a clearer organization of the methodological landscape, we categorize existing VQA methods based on their backbone architectures: Statistical-based, CNN-based, Transformer-based, and CLIP-based methods.

\subsubsection{Statistical-based VQA}

Statistical-based methods rely on handcrafted features combined with traditional regression models, offering computational efficiency and interpretability.

\textbf{VMAF}\cite{VMAF} combines spatial and temporal quality-aware features using a Support Vector Regression (SVR) model to predict video quality, incorporating metrics such as Detail Loss Metric (DLM), Visual Information Fidelity (VIF), and Temporal Information (TI). Enhancements like SpatioTemporal VMAF and Ensemble VMAF address its initial limitations in temporal quality measurement, improving accuracy and generalization. \textbf{Temporal Pooling}\cite{EPooling} refines quality predictions by mapping input features to frame-level scores and fusing them through a regression process that integrates diverse pooling methods, including Hysteresis. Both approaches leverage temporal features to handle varying motion and quality dynamics effectively, demonstrating robust performance across datasets and broad applicability in real-time video quality assessment.

\subsubsection{CNN-based VQA}

CNN-based methods leverage convolutional neural networks\cite{10.1145/3065386} to automatically learn spatial and temporal quality features from video data, enabling end-to-end quality prediction. We further categorize CNN-based methods based on their architectural innovations and feature extraction strategies.

\textit{Temporal modeling methods.} These methods focus on capturing temporal dependencies and long-range quality variations across video frames. \textbf{VSFA}\cite{VSFA} integrates human visual system knowledge with a pre-trained CNN for content-aware feature extraction and a GRU \cite{chung2014empiricalevaluationgatedrecurrent} for modeling long-term dependencies, employing a temporal pooling layer to align predictions with human perception. \textbf{MDTVSFA}\cite{MDTVSFA} employs multi-stage frameworks with relative quality assessors and nonlinear mapping for consistent rankings and perceptual nonlinearity handling. \textbf{GSTVQA}\cite{GSTVQA} utilizes multi-scale feature extraction and attention mechanisms to capture both global and local quality information.

\textit{Feature fusion methods.} These methods combine multiple feature sources or leverage transfer learning to enhance quality prediction robustness. \textbf{VIDEVAL}\cite{VIDEVAL} enhances efficiency by combining features from high-performing blind VQA models, selecting 60 key features and incorporating transfer learning with ResNet-50 for robust assessments in UGC-VQA contexts. \textbf{RAPIQUE}\cite{RAPIQUE} combines spatial and temporal scene statistics with deep learning features for efficient and accurate assessments, particularly for motion-intensive content. \textbf{CompressedVQA}\cite{FR_NR_VQA} integrates CNN-based feature extraction with temporal pooling strategies to capture complex quality relationships, achieving state-of-the-art performance on compressed video datasets.

\textit{Multi-scale analysis methods.} These methods extract features at multiple spatial or temporal scales to capture both local distortions and global quality impressions. \textbf{Patch-VQ}\cite{Patch_VQ} adopts a local-to-global architecture with 2D and 3D feature extraction to capture distortions, generating space-time quality maps for detailed insights into video degradation. \textbf{SimpleVQA}\cite{Pro.} leverages multi-scale fusion strategies and pretrained action recognition networks for motion feature extraction. \textbf{HVS-5M}\cite{HVS_5M} integrates five representative Human Visual System (HVS) characteristics, including saliency maps, content dependency, and temporal hysteresis, for comprehensive evaluations. \textbf{Xu et al.}\cite{MultiLevelFusion} decomposes videos into frame, segment, and video levels, employing spatial-temporal data augmentation, multi-level feature fusion, and adaptive rank-aware loss to address hard samples with subtle quality distinctions, achieving strong performance in the \textbf{NTIRE'24 S-UGC Challenge} \cite{NTIRE2024_SUGC}.

\textit{Quality disentanglement methods.} These methods decompose video quality into multiple independent dimensions for more interpretable assessment. \textbf{DOVER}\cite{wu_exploring_2023-1} disentangles aesthetic and technical evaluations, supported by the extensive DIVIDE-3k dataset, providing nuanced insights into human perceptual mechanisms. \textbf{MD-VQA}\cite{MD_VQA} focuses on UGC live videos with multi-dimensional approaches combining semantic, distortion, and motion features. \textbf{Light-VQA}\cite{Light_VQA} targets low-light video assessments using handcrafted and deep learning-based features. \textbf{Ada-DQA}\cite{Ada_DQA} introduces an adaptive framework with pretrained models and a dynamic Quality-aware Acquisition Module (QAM) to optimize lightweight VQA models. \textbf{BVQA}\cite{BVQA} employs a modular design with spatial and temporal rectifiers to refine quality predictions, ensuring precision and extensibility.

\subsubsection{Transformer-based VQA}

Transformer-based methods utilize self-attention mechanisms\cite{vaswani2017attention} to capture long-range spatiotemporal dependencies, enabling more effective modeling of complex quality patterns across video sequences. We organize these methods based on their attention mechanisms and architectural designs.

\textit{Attention-based methods.} These methods design specialized attention mechanisms to model spatial and temporal quality relationships. \textbf{StarVQA}\cite{StarVQA} employs a space-time attention network to capture long-range spatiotemporal dependencies, utilizing a vectorized regression loss and an alternating attention mechanism to process high-resolution videos effectively. \textbf{FANet}\cite{FANet}, with its modified Video Swin Transformer backbone, introduces Gated Relative Position Biases (GRPB) and an Intra-Patch Non-Linear Regression (IP-NLR) head to deliver efficient and precise quality predictions. \textbf{VQT}\cite{VQT} introduces Sparse Temporal Attention (STA) and a Multi-Pathway Temporal Network (MPTN) to efficiently process keyframes and capture multi-distortion characteristics. \textbf{SSL-VQA}\cite{SSL_VQA} uses a self-supervised spatio-temporal framework to operate effectively with limited labeled data through statistical contrastive learning. \textbf{MVQA}\cite{MVQA_Mamba} is the first to apply state-space models (Mamba) for VQA, achieving linear complexity with sequence length. 

\textit{Hierarchical methods.} These methods adopt multi-level architectures to capture quality information at different granularities. \textbf{Zoom-VQA}\cite{Zoom_VQA} adopts a hierarchical approach with patch, frame, and clip-level feature extraction to comprehensively analyze video quality, from both image and video quality assessment branches. \textbf{SB-VQA}\cite{SB_VQA} uses a stack-based architecture fine-tuned on the PGCVQ dataset to evaluate professional video content. \textbf{COVER}\cite{COVER} enhances quality evaluation by fusing technical, aesthetic, and semantic features through a Simplified Cross-Gating Block (SCGB), processing videos through three parallel branches. \textbf{Unified-VQA}\cite{UnifiedVQA} proposes a Diagnostic Mixture-of-Experts (MoE) framework that employs multiple perceptual experts dedicated to distinct domains (HD, UHD, HDR, HFR). 

\textit{Prior-guided methods.} These methods incorporate external knowledge or priors to guide quality prediction. \textbf{Light-VQA+}\cite{Light_VQA_plus} integrates vision-language guidance and trainable attention weights to improve evaluation for low-light and over-exposed videos. \textbf{PTM-VQA}\cite{PTM_VQA} employs pretrained models and a dual-constraint loss to ensure consistency and separability in its quality-aware latent space. \textbf{PriorFormer}\cite{Priorformer} combines content and distortion priors with GRU\cite{chung2014empiricalevaluationgatedrecurrent} to capture semantic and distortion features dynamically over time. \textbf{ReLaX-VQA}\cite{ReLaX_VQA} uses spatio-temporal fragment sampling and optical flow techniques to address UGC video complexities. 

\subsubsection{CLIP-based VQA}

CLIP-based methods leverage vision-language pretraining\cite{radford2021learning} to incorporate semantic understanding into quality assessment, enabling zero-shot or few-shot quality prediction and better alignment with human quality descriptions. We categorize these methods based on their utilization of vision-language representations.

\textit{Semantic-driven methods.} These methods utilize CLIP's semantic representations to assess quality without explicit human annotations. \textbf{BVQI}\cite{BVQI} integrates the Semantic Affinity Quality Index (SAQI) and its localized variant to evaluate video quality based on semantic content. By incorporating Spatial and Temporal Naturalness Indices and optimizing text prompts, BVQI captures both aesthetic and authentic distortions effectively. \textbf{MaxVQA}\cite{MaxVQA} combines CLIP visual features with FAST-VQA features to assess nuanced quality dimensions using the Maxwell database with over two million human opinions on 13 quality factors.

\textit{UGC-focused methods.} These methods are designed for user-generated and short-form video content. \textbf{KSVQE}\cite{KSVQE} addresses short-form UGC challenges by leveraging the KVQ database and incorporating Quality-Aware Region Selection (QRS) and Content-Adaptive Modulation (CaM) modules. \textbf{RQ-VQA}\cite{RQ_VQA}, designed for social media videos, integrates features from BIQA and BVQA models with trainable spatial and fixed temporal quality modules for precise predictions.

\textit{AIGC-oriented methods.} These methods focus on evaluating AI-generated video content. \textbf{UGVQ}\cite{UGVQ} evaluates spatial quality, temporal quality, and text-to-video alignment by combining CLIP for feature extraction and SlowFast for motion representation. \textbf{NTIRE'24 AIGC-QA Challenge}\cite{NTIRE2024_AIGCQA} further advances this direction using the AIGIQA-20K and T2VQA-DB datasets. \textbf{CLIPVQA}\cite{CLIPVQA} proposes an efficient CLIP-based Transformer that transforms MOS into quality language descriptions, aggregating video content with language features via cross-attention for video-language quality representations.

\subsubsection{LMM-based VQA}

Large Multimodal Model (LMM)-based methods leverage the powerful visual understanding and reasoning capabilities of foundation models to perform video quality assessment through instruction tuning and question-answering paradigms.

\textit{Instruction-tuned method.} These methods reformulate quality regression as a question-answering task and fine-tune LMMs on VQA datasets. \textbf{Q-Align}\cite{QAlign} proposes to teach LMMs with discrete text-defined rating levels instead of numerical scores, emulating the human subjective rating process. 
\textbf{LMM-VQA}\cite{LMM_VQA} introduces a spatiotemporal visual modeling strategy that designs quality-aware feature extraction through a dedicated spatiotemporal vision encoder and projector, achieving an average 5\% improvement in generalization over existing methods. \textbf{VQA$^2$}\cite{VQA2} presents the first visual question answering instruction dataset for video quality assessment containing 157,755 question-answer pairs, and develops the VQA$^2$ series models that interleave visual and motion tokens for spatiotemporal quality perception.

\textit{Efficient LMM-based method.} These methods aim to reduce the computational overhead of LMM-based VQA while maintaining performance. \textbf{Q-CLIP}\cite{QCLIP_2025} proposes a Symmetric Cross-Modal Adapter (SCMA) that enables efficient adaptation of vision-language models with minimal trainable parameters, introducing learnable five-level quality prompts to guide fine-grained quality perception. The method investigates frame-difference-based sampling strategies for improved cross-dataset generalization.

\subsubsection{Technical Recommendations}

Based on our comprehensive review, we provide the following guidance for practitioners and researchers selecting VQA methods:
For \textbf{real-time applications} with limited computational resources, Statistical-based methods like VMAF~\cite{VMAF} remain strong choices due to their efficiency and interpretability, particularly for streaming quality monitoring. When \textbf{computational resources are available} and high accuracy is prioritized, CNN-based methods offer a good balance between performance and complexity, with temporal modeling approaches (VSFA~\cite{VSFA}, MDTVSFA~\cite{MDTVSFA}) suitable for general UGC content and multi-scale methods (Patch-VQ~\cite{Patch_VQ}, Xu et al.~\cite{MultiLevelFusion}) preferred when fine-grained quality localization is needed.

For applications requiring \textbf{long-range temporal understanding} or handling \textbf{high-resolution videos}, Transformer-based methods are recommended. Attention-based approaches (StarVQA~\cite{StarVQA}, FANet~\cite{FANet}) excel at capturing global dependencies, while hierarchical methods (Zoom-VQA~\cite{Zoom_VQA}, COVER~\cite{COVER}) are better suited for multi-dimensional quality analysis. Prior-guided methods (PTM-VQA~\cite{PTM_VQA}, PriorFormer~\cite{Priorformer}) are particularly effective when domain-specific knowledge can be incorporated.
For \textbf{AI-generated video evaluation} or scenarios requiring \textbf{semantic-aware assessment}, CLIP-based methods provide superior performance by leveraging vision-language alignment. AIGC-oriented methods (UGVQ~\cite{UGVQ}, CLIPVQA~\cite{CLIPVQA}) are specifically designed for text-to-video generation evaluation, while UGC-focused methods (KSVQE~\cite{KSVQE}, RQ-VQA~\cite{RQ_VQA}) address the unique challenges of short-form social media content.
For applications requiring \textbf{explainable quality assessment} or \textbf{natural language quality descriptions}, LMM-based methods represent the emerging frontier. Q-Align and VQA$^2$ excel at providing human-aligned quality ratings through instruction tuning, while LMM-VQA offers strong spatiotemporal understanding. However, LMM-based methods typically require significant computational resources; for resource-constrained scenarios, efficient alternatives like Q-CLIP or MVQA provide better trade-offs between performance and computational cost.


\subsection{Evaluation Metrics}
The algorithms mentioned in previous sections are typically evaluated based on the correlation between subjective and objective ratings. Five frequently used metrics highlight various facets of VQA model performance: SRCC, KRCC, and PLCC measure prediction monotonicity through correlation, while RMSE and MAE measure prediction accuracy through error computation. Better performance is indicated by higher (closer to 1) correlation values and lower (closer to 0) error values.

\textbf{Spearman Rank-order Correlation Coefficient (SRCC)} is a non-parametric measure evaluating the monotonic relationship between two ranked variables. Unlike Pearson's correlation which assesses linear relationships, SRCC is useful when data do not meet normality assumptions or when relationships are nonlinear. It works by ranking the data points and then applying correlation to these ranks:
$$
\mathrm{SRCC}=1-\frac{6 \sum_{i=1}^N d_i^2}{N\left(N^2-1\right)}
$$
where $N$ is the number of test videos and $d_i$ is the rank difference between subjective and objective scores for the $i$-th video. SRCC is commonly used when ordinal data or nonlinear associations are present, making it a versatile tool in VQA evaluation.

\textbf{Kendall Rank-order Correlation Coefficient (KRCC)} measures ordinal association based on concordant and discordant pairs of observations. A concordant pair occurs when the order of ranks for both variables is consistent, while a discordant pair occurs when the order is reversed:
$$
\mathrm{KRCC}=\frac{N_c-N_d}{\frac{1}{2} N(N-1)}
$$
where $N_c$ and $N_d$ are the numbers of concordant and discordant pairs, respectively. KRCC is particularly useful for small datasets or when many tied ranks exist, providing a robust method for assessing ordinal relationships.

\textbf{Pearson Linear Correlation Coefficient (PLCC)} quantifies the strength and direction of the linear relationship between two continuous variables. The coefficient ranges from -1 to +1, where +1 indicates a perfect positive linear relationship, -1 indicates a perfect negative linear relationship, and 0 suggests no linear relationship:
$$
\operatorname{PLCC}=\frac{\sum_i^N\left(q_i-\bar{q}\right) \cdot\left(o_i-\bar{o}\right)}{\sqrt{\sum_i^N\left(q_i-\bar{q}\right)^2 \cdot\left(o_i-\bar{o}\right)^2}}
$$
where $o_i$ and $q_i$ are the subjective opinion score and nonlinear-fitted objective score for the $i$-th video, and $\bar{o}$ and $\bar{q}$ are their mean values. PLCC is sensitive to outliers and should be used when a linear relationship is expected.

\textbf{Root Mean Square Error (RMSE)} measures the magnitude of prediction errors by calculating the square root of the average squared differences between predicted and observed values:
$$
\mathrm{RMSE}=\sqrt{\frac{1}{N} \sum_i^N\left(q_i-o_i\right)^2}
$$
where $q_i$ represents the observed values and $o_i$ the predicted values. RMSE is sensitive to large errors due to the squaring process, making it useful when large deviations are of greater concern.

\textbf{Mean Absolute Error (MAE)} measures the average magnitude of errors without considering their direction:
$$
\mathrm{MAE}=\frac{1}{N} \sum_i^N\left|q_i-o_i\right|
$$
Unlike RMSE, MAE does not give extra weight to large errors, making it less sensitive to outliers and more interpretable when all errors should be treated with equal importance.

%% file: content/5-tv_align.tex
\section{Alignment with human Instructions}
\label{sec_instruction}

Generating videos from specified human instructions has long been an important topic in video generation. It is also an essential aspect that many current studies of AI-generated videos strive to improve. 
Over the past decade, numerous benchmark datasets and evaluation methods have been proposed to assess the alignment of generated videos with human instructions (i.e., text, audio).
This section introduces the representative benchmark datasets and evaluation methods.

\subsection{Benchmark Datasets}

\begin{figure}
    \centering
    \includegraphics[width=0.85\textwidth]{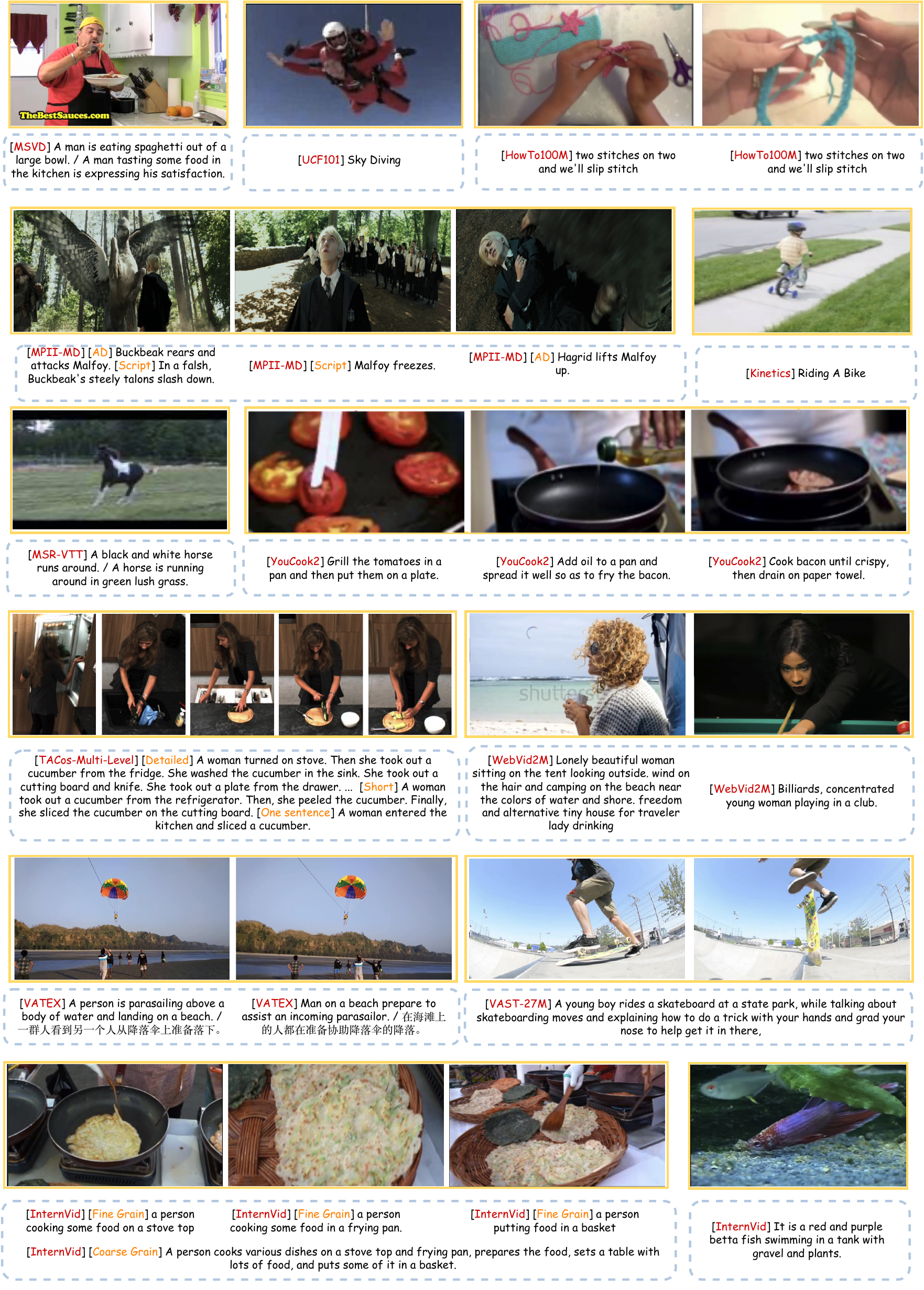}
    \caption{Exemplar Cases from Alignment with Human Instructions Benchmark Datasets. Here we intercept one or several frames of the video clip as a representation of the video within each dataset, along with the different corresponding textual representations.}
    \label{fig:inst_dataset_case}
    \vspace{-2mm}
\end{figure}

\begin{table}[!t]
\centering
\caption{Summary of the mentioned video-human instruction alignment benchmark datasets.}
\scalebox{0.68}{\begin{tabular}{l l l l l l l l l c}
\toprule
\textbf{Dataset} & \textbf{Year} & \textbf{Instruction Type} & \textbf{Domain} & \#\textbf{Video} & \#\textbf{Video Clips} & \#\textbf{Sentence} & \textbf{Duration(hrs)} & \textbf{Resolution} & \textbf{Link} \\
\midrule
MSVD \cite{chen-dolan-2011-collecting} & 2011 & text & multi-category & - & 1,970 & 70,028 & 5.3 & - & \href{https://www.cs.utexas.edu/~ml/clamp/videoDescription/}{\includegraphics[width=0.03\textwidth]{figures/github_logo.png}} \\
UCF101 \cite{soomro2012ucf101} & 2012 & text & 101 action classes & - & 13,320 & - & 27 & 320p & 
\href{https://www.crcv.ucf.edu/data/UCF101.php}{\includegraphics[width=0.03\textwidth]{figures/github_logo.png}} \\
MPII-MD \cite{rohrbach2015dataset} & 2015 & text & movie & 94 & 68,337 & 68,337 & 73.6 & 720p &
\href{https://www.mpi-inf.mpg.de/departments/computer-vision-and-machine-learning/research/vision-and-language/mpii-movie-description-dataset}{\includegraphics[width=0.03\textwidth]{figures/github_logo.png}} \\
MSR-VTT \cite{xu2016msr} & 2016 & text & multi-category & 7,180 & 10,000 & 200,000 & 41.2 & 240p &
\href{https://www.microsoft.com/en-us/research/publication/msr-vtt-a-large-video-description-dataset-for-bridging-video-and-language/}{\includegraphics[width=0.03\textwidth]{figures/github_logo.png}} \\
Kinetics \cite{kay2017kinetics} & 2017 & text & 400 action classes & - & 306,245 & - & $\sim$ 850 & 340p/128p & 
\href{https://github.com/cvdfoundation/kinetics-dataset}{\includegraphics[width=0.03\textwidth]{figures/github_logo.png}} \\
YouCook2 \cite{zhou2018towards} & 2018 & text & cooking & 2,000 & 14,000 & 14,000 & 176 & - &
\href{http://youcook2.eecs.umich.edu}{\includegraphics[width=0.03\textwidth]{figures/github_logo.png}} \\
TACos-Multi-Level \cite{rohrbach2014coherent} & 2014 & text & cooking & 273 & 14,105 & 52,593 & 176 & - & 
\href{https://www.mpi-inf.mpg.de/departments/computer-vision-and-machine-learning/research/vision-and-language/tacos-multi-level-corpus}{\includegraphics[width=0.03\textwidth]{figures/github_logo.png}} \\
HowTo100M \cite{miech2019howto100m} & 2019 & text & instruction & 1.22M & 136M & 136M & 134,472 & 240p & 
\href{https://www.di.ens.fr/willow/research/howto100m/}{\includegraphics[width=0.03\textwidth]{figures/github_logo.png}} \\
VATEX \cite{wang2019vatex} & 2019 & text & open & 41,269 & 41,269 & 825,380 & $\sim$ 115 & - &
\href{https://eric-xw.github.io/vatex-website/about.html}{\includegraphics[width=0.03\textwidth]{figures/github_logo.png}} \\
Webvid-2M \cite{bain2021frozen} & 2021 & text & instruction & - & 2.5M & 2.5M & 13K & 360p &
\href{https://github.com/m-bain/webvid}{\includegraphics[width=0.03\textwidth]{figures/github_logo.png}} \\
InternVid \cite{wang2023internvid} & 2023 & text & open & 7.1M & 234M & 234M & 760.3 & 720p &
\href{https://github.com/OpenGVLab/InternVideo/tree/main/Data/InternVid}{\includegraphics[width=0.03\textwidth]{figures/github_logo.png}} \\
Panda-70M \cite{chen2024panda} & 2024 & text & open & 3.8M & 70.8M & 70.8M & 166.8K & 720p &
\href{https://snap-research.github.io/Panda-70M/}{\includegraphics[width=0.03\textwidth]{figures/github_logo.png}} \\
VAST-27M \cite{chen2024vast} & 2024 & text, audio & open & 3.3M & 27M & 297M & 75K & 720p &
\href{https://github.com/TXH-mercury/VAST}{\includegraphics[width=0.03\textwidth]{figures/github_logo.png}} \\

\bottomrule

\end{tabular}
}
\label{tab:inst_dataset}
\end{table}
In this section, we introduce several representative benchmark datasets widely used before and during the era of AIGV to assess the alignment between video and human instruction.
Each benchmark dataset contains thousands to millions of video clips, along with the corresponding descriptions that are either manually annotated or generated by multimodal models. Table \ref{tab:inst_dataset} provides an overview of the key details for each benchmark dataset mentioned, while Figure  \ref{fig:inst_dataset_case} illustrates an exemplar case from each dataset. 
Following the categories of textual instructions, we classify the datasets into the following three groups.

\subsubsection{Short label Dataset. } A substantial number of early datasets use short labels (i.e., 1-4 words) as the textual instructions. \textbf{UCF101} \cite{soomro2012ucf101} is a popular video action recognition dataset and contains 101 action classes. These short labels are divided into five action types, addressing challenges such as poor lighting and severe camera motion, with fixed resolutions and fps. 
\textbf{Kinetics} \cite{kay2017kinetics} is a dataset with a broader scope. It offers 400 action classes with diverse second-level video clips sourced from YouTube. Successors Kinetics-600 \cite{carreira2018shortnotekinetics600} and Kinetics-700 \cite{carreira2022shortnotekinetics700human} expand the labels from Kinetics-400, providing more classes (i.e., 600 and 700, respectively) while ensuring unique sources for each clip.

\subsubsection{One-sentence Narrative Dataset. } Beyond the use of short labels for textual instruction, rapid progress in video processing technologies has led to a growing number of datasets centered on narrative sentences proposed. \textbf{MSVD} \cite{chen-dolan-2011-collecting} is a widely-used benchmark dataset for video description. These descriptions, collected via Amazon Mechanical Turk, summarize unambiguous human events in muted video segments. \textbf{MPII-MD} \cite{rohrbach2015dataset}: MPII-MD evaluates video-text alignment in the movie domain. By combining audio descriptions for visually impaired audiences with script mining, it offers a unique long-format benchmark for aligning text with extended video content. Compared with previous work, \textbf{MSR-VTT} \cite{xu2016msr} provides a relatively large-scale benchmark for video-to-text translation across 20 categories, including music, sports, and gaming, and offers comprehensive, diverse video-text pairs for early benchmarking. 
Some datasets focus on specific domains or even have unique structures. \textbf{YouCook2} \cite{zhou2018towards} uses cooking videos from YouTube as video content. It addresses challenges like fast camera motion and provides detailed annotations for global cuisines and cooking methods. \textbf{TACos-Multi-Level} \cite{rohrbach2014coherent} provides multi-grain descriptions (single sentence, 3-5 sentences, and 15 sentences) for cooking videos. It uses intermediate semantic representations to align textual descriptions with video snippets, highlighting how description compression varies with length.

\subsubsection{Large-scale Multi-domain Dataset. } With the evolution of Automatic Speech Recognition (ASR) and Large Language Models (LLMs), generating textual instructions for video content has gradually become an automated task, thereby significantly increasing dataset scale and domain breadth without the burden of manual effort. \textbf{HowTo100M} \cite{miech2019howto100m} is a dataset of instructional YouTube video clips covering 23,000 tasks, such as cooking, gardening, and crafting. By leveraging narration subtitles generated by YouTube ASR, it facilitates scalable and resource-efficient dataset creation for instructional video understanding. \textbf{VATEX} \cite{wang2019vatex} advances multilingual video-text alignment with bilingual captions in English and Chinese. Built on Kinetics-600, it provides hundreds of thousands of English-Chinese parallel pairs, supporting tasks like video-guided machine translation and multilingual text alignment. \textbf{WebVid-2M} \cite{bain2021frozen} is a dataset comprising millions of video-text pairs collected via web scraping and filtering, providing diverse open-domain descriptions. Captions range from poetic to succinct, enhancing their utility for broad video-text alignment tasks. \textbf{InternVid} \cite{wang2023internvid} is a dataset comprising over 7M videos and 234M multilingual clip captions. It uses models such as Tag2Text \cite{huang2023tag2text} and BLIP2 \cite{li2023blip}, it produces fine- and coarse-grained descriptions, supporting diverse scenarios and robust multimodal understanding. \textbf{Panda-70M} \cite{chen2024panda} is a dataset constructed from HD-VILA-100M videos \cite{xue2022advancinghighresolutionvideolanguagerepresentation}. It uses semantic splitting and cross-modality teacher models to create 70M fine-grained video-text pairs, emphasizing alignment accuracy and diversity and supporting detailed video semantics analysis. \textbf{VAST-27M} \cite{chen2024vast} is a dataset that integrates vision, audio, and subtitle captions for 27M clips across 15 categories. It uses LLMs such as Vicuna-13b \cite{zheng2023judgingllmasajudgemtbenchchatbot} for omni-modal captions, and supports tasks beyond text, including retrieval, captioning, and multimodal video analysis.

\begin{figure}[!t]
    \centering
    \includegraphics[width=0.70\linewidth]{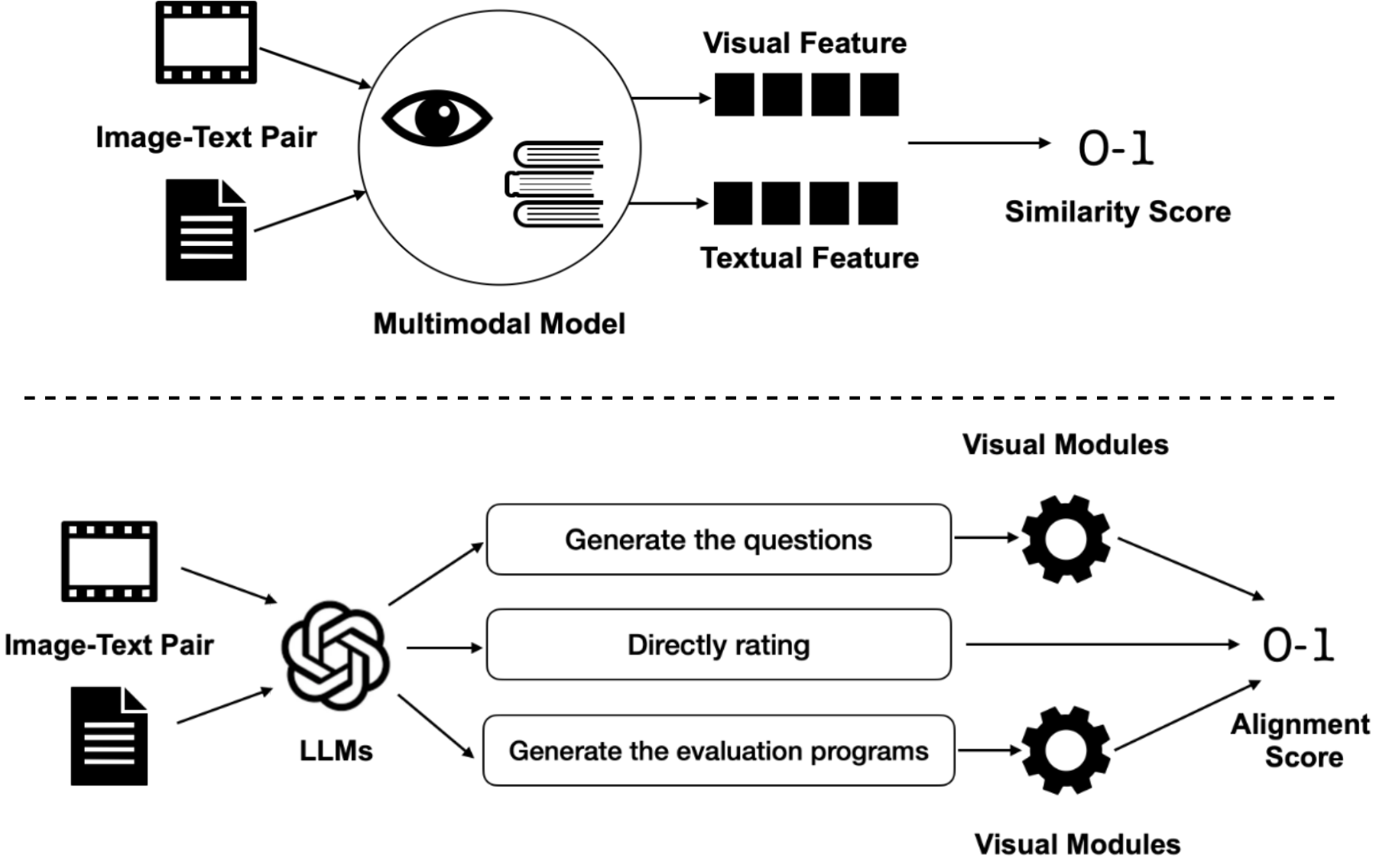}
    \caption{Compared with the previous multimodal evaluation backbone (top), the emergence of LLMs helps visual-textual alignment evaluation tasks evolve towards diversity, making the evaluation process more interpretable (bottom).}
    \label{fig:03}
\end{figure}

\begin{figure*}[t]
\centering
\small
\scalebox{0.80}{
\begin{forest}
for tree={
    grow'=0,
    anchor=west,
    parent anchor=east,
    child anchor=west,
    edge={gray!70, line width=0.9pt},
    rounded corners=6pt,
    l sep=10mm,
    s sep=5mm,
    inner xsep=7pt,
    inner ysep=5pt,
    minimum height=8mm,
    edge path={
        \noexpand\path[\forestoption{edge}]
        (!u.east) -- +(8pt,0) |- (.west)\forestoption{edge label};
    },
    font=\normalsize,
    draw=none,
},
root/.style={
    draw=black!70,
    fill=gray!15,
    rounded corners=8pt,
    font=\bfseries,
},
multi/.style={
    draw=red!70,
    fill=red!10,
    font=\bfseries,
},
multicite/.style={
    draw=red!70,
    fill=red!5,
},
llm/.style={
    draw=green!60!black,
    fill=green!10,
    font=\bfseries,
},
llmcite/.style={
    draw=green!60!black,
    fill=green!5,
},
mid/.style={
    font=\normalsize,
},
leaf/.style={
    font=\normalsize,
},
[Alignment with Human Instructions, root
    [Multimodal-based, multi
        [Stack Cross Attention, multi
            [{{\cite{jiang2019tiger}}}, multicite, leaf]
        ]
        [ViLBERT, multi
            [{{\cite{lu2019vilbert}, \cite{inan2021cosmic}}}, multicite, leaf]
        ]
        [CLIP/BLIP, multi
            [{{\cite{hessel2022clipscore}, \cite{kim2022mutual}, \cite{kirstain2023pick}}}, multicite, leaf]
        ]
        [Human Preferences, multi
            [{{\cite{ji2024tltscore}, \cite{xu2024imagereward}}}, multicite, leaf]
        ]
    ]
    [LLM-based, llm
        [Directly Reasoning the Consistency, llm
            [{{\cite{zhang2023gpt}, \cite{lu2024llmscore}, \cite{ku2023viescore}}}, llmcite, leaf]
        ]
        [Question Generation and Answering, llm
            [{{\cite{hu2023tifa}, \cite{yarom_what_nodate}, \cite{cho2023davidsonian}}}, llmcite, leaf]
        ]
        [Interpretable Visual Programming, llm
            [{{\cite{cho2024visual}}}, llmcite, leaf]
        ]
    ]
]
\end{forest}

}

\caption{Taxonomy of Alignment with Human Instructions Methods.}
\label{fig:alignment_human_instructions_mindmap}
\end{figure*}

\subsection{Evaluation Methods}

In this section, we introduce several recent representative video-text alignment evaluation models. 
Different from the traditional evaluation metrics \cite{anderson2016spice, vedantam2015ciderconsensusbasedimagedescription, banerjee-lavie-2005-meteor, lin2004rouge, Papineni2002BleuAM}, which based solely on assessing the accuracy and fluency of generated video captions, the following studies consider both visual and textual elements, which could better align with the era of AI video generation.
Constrained by the computation bottleneck of end-to-end processing of long videos, the current mainstream methods for modeling evaluation are mainly frame-based. That is, they use images as visual representations to measure the alignment between key video frames and human instructions. 
As shown in Figure \ref{fig:03}, early modeling evaluation studies primarily used the multimodal pre-trained models as their core backbone. However, with recent advances, there has been a shift toward large language models (LLMs), which have become increasingly prominent in evaluation methods. The taxonomy of alignment with the human instructions method in Figure \ref{fig:alignment_human_instructions_mindmap}.

\subsubsection{Multimodal-based Methods} 
The advancement of cross-modal pretraining has encouraged the development of multimodal-based evaluation methods that contrast the embedding similarity between images and text within a shared latent space.
\textbf{TIGEr} \cite{jiang2019tiger} evaluates image-text alignment by addressing limitations of traditional text-based metrics, which rely solely on textual matching between reference and generated captions. It uses a pre-trained SCAN model \cite{lee2018stacked} to first compute grounding scores for image-text pairs, then calculates the similarity between grounding vectors for generated and reference captions based on Region Rank Similarity (RRS) and Weight Distribution Similarity (WDS), which are averaged to produce the final TIGEr score. 
\textbf{ViLBERTScore} \cite{lee2020vilbertscore} improves image-caption evaluation by using contextual embeddings from the ViLBERT model \cite{lu2019vilbert}, which integrates both visual and textual data. Region-level features of images are extracted via object detection \cite{he2017mask}, and pairs of generated and reference captions are embedded with their corresponding image features. Cosine similarity between embeddings is computed using greedy token matching, reflecting alignment between captions and visual content.
\textbf{COSMic} \cite{inan2021cosmic} introduces the first discourse-aware metric for image caption evaluation, addressing challenges in assessing captions with varying goals and perspectives. Using the COIN dataset, which categorizes captions based on coherence labels like \textit{Meta}, \textit{Visible}, \textit{Subjective}, and \textit{Story}, COSMic is trained to accommodate diverse discourse goals. It incorporates both ViLBERT-based \cite{lu2019vilbert} and Vanilla models to assess coherence and alignment between captions and image content.

\textbf{CLIP-based Methods.} CLIP \cite{radford2021learning} is a widely adopted pre-trained multi-modal model. Its alignment-friendly encoding of textual and visual modalities has laid the foundation for various evaluation studies.
\textbf{CLIPScore} \cite{hu2023tifa} leverages the CLIP model to compute alignment scores between images and captions. It calculates cosine similarity between visual and textual embeddings without requiring references, enabling scalable evaluation. 
\textbf{PickScore} \cite{kirstain2023pick} builds on the CLIPScore framework by incorporating user preferences captured through the Pick-a-Pic dataset, which includes over 500,000 examples of text prompts and corresponding generated images. By training on human-preferred image pairs using reinforcement learning, PickScore aligns evaluations with real-world user expectations. It computes similarity scores using inner products of CLIP-based embeddings and fine-tunes the evaluation model with an InstructGPT-inspired objective \cite{ouyang2022training}.
\textbf{Mutual Information Divergence (MID)} \cite{kim2022mutual} evaluates image-text alignment by applying mutual information concepts with CLIP-based pre-trained encoders. It uses negative Gaussian cross-mutual information to calculate alignment, bridging the gap between visual reasoning and textual evaluation. MID derives point-wise mutual information for pairwise evaluation and computes the expectation of PMI as the final score.

\textbf{Human Preferences Alignment Methods.} In addition to standard similarity score calculations, some studies have introduced additional concepts to align evaluation biases more closely with human preferences. \textbf{ImageReward} \cite{xu2024imagereward} employs BLIP \cite{li2023blip} as its backbone model, using human-annotated preferences to train a reward model for evaluating text-to-image alignment. The dataset, derived from DiffusionDB \cite{wang2022diffusiondb}, includes text prompts paired with multiple images, rated and ranked by human annotators. The model is trained to minimize discrepancies between predicted and actual preferences using reinforcement learning. 
\textbf{TITScore} \cite{ji2024tltscore} addresses the long-tailed effect in text-to-image evaluation by integrating symbolic-level understanding with neural-level reasoning. It employs a mixture-of-experts (MOE) approach, combining curated prompts with visual reasoning models like segmentation and detection \cite{kirillov2023segment, oquab2023dinov2}. By enhancing prompt embeddings with aspect-specific token decompositions, TITScore evaluates text-to-image alignment with greater precision.

\subsubsection{LLM-based Methods}

Large Language Models (LLMs) have emerged as powerful tools for evaluating text-to-image alignment and video quality, leveraging their multimodal reasoning capabilities to address the complex nature of vision-language tasks. \textbf{GPT-4V Eval} \cite{zhang2023gpt} demonstrates the potential of GPT-4V \cite{2023GPT4VisionSC} as an evaluator for vision-language tasks, introducing \textit{single-answer grading} and \textit{pairwise comparison} pipelines to assess tasks such as image-to-text captioning and text-guided image editing. 
\textbf{LLMScore} \cite{lu2024llmscore} enhances image-text alignment evaluation by leveraging multi-granularity visual descriptors processed by models such as BLIPv2 \cite{li2023blip}. Using GPT-4 \cite{openai2023gpt4} as the evaluator, it combines reasoning and compositionality to generate detailed alignment scores that mimic human evaluation processes. \textbf{VIEScore} \cite{ku2023viescore} also uses LLMs as the backbone (e.g., GPT-4o \cite{achiam2023gpt}), but incorporates task-specific rating instructions to measure semantic consistency and perceptual quality (SC and PQ), and derives final scores by combining SC and PQ. \textbf{TIFA} \cite{hu2023tifa} utilizes question generation and answering (QG/A) frameworks to evaluate text-to-image alignment. TIFA assesses faithfulness by generating binary QA pairs from text prompts and verifying alignment through VQA models. 
\textbf{VQ2} \cite{yarom_what_nodate} extends the Q/A approach by using candidate answers to generate yes-no QA pairs, deriving alignment scores for text-image pairs. 
\textbf{Davidsonian Scene Graph (DSG)} \cite{cho2023davidsonian} refines QG/A methodologies by ensuring questions are atomic and unique, leveraging directed acyclic graphs (DAGs) to represent semantic dependencies. It uses high-performance LLMs such as GPT-4 and PaLM2 \cite{anil2023palm} for question generation and validates the generated questions using VQA models.
Meanwhile, \textbf{VPEVAL} \cite{cho2024visual} adopts an innovative visual programming approach, generating evaluation programs via LLMs for both skill-based and open-ended tasks. By invoking diverse visual modules such as Grounding DINO \cite{liu2023grounding} and BLIP-2 \cite{li2023blip}, VPEVAL evaluates alignment across multiple skills, offering a dynamic and interpretable framework.

Together, these LLM-based methods illustrate the transformative role of multimodal reasoning and adaptive frameworks in enhancing the evaluation of vision-language tasks, providing robust, explainable, and scalable approaches for aligning textual and visual content.

Multimodal-based methods typically use multimodal models to project visual and textual descriptions into a shared latent space, where alignment is quantified via embedding similarity. This encoding-based judgment maintains a degree of consistency determined by the backbone models.
Meanwhile, LLM-based methods rely on reasoning-driven evaluation. While heavily influenced by generation quality, these methods provide enhanced interpretability and broader scalability for diverse evaluation dimensions. 
Therefore, multimodal-based approaches facilitate computational efficiency and standardized outcomes, as most backbone models are relatively compact in scale. In contrast, LLM-based methods are more compute-intensive, resulting in higher latency and producing more interpretable, reasoning-based evaluation feedback.

%% file: content/6-multi_asp.tex

%% file: content/8-future.tex
\section{Future Prospects}
\label{sec_future}

Despite rapid progress, current research in AI-Generated Video Evaluation (AIGVE) still faces substantial limitations that hinder its reliability, interpretability, and real-world deployment. Rather than viewing future development solely as opportunities, it is important to critically examine these limitations and outline potential research directions to address them.

\textbf{Leveraging Vision-Language Models Beyond Surface Alignment}.  
Existing AIGVE methods often rely on handcrafted metrics or task-specific models that struggle to capture high-level semantics, reasoning, and cross-modal dependencies. They typically evaluate perceptual fidelity or instruction alignment in isolation, leading to fragmented assessments and limited generalization across scenarios. Furthermore, many current models lack contextual understanding of temporal causality, commonsense reasoning, or long-range narrative coherence.

Vision-Language Models (VLMs), such as Qwen \cite{qwen}, LLaVA \cite{liu2023llava}, and Chameleon \cite{chameleon2024}, offer promising capabilities for addressing these limitations through unified multimodal reasoning. Their transformer-based architectures \cite{vaswani2017attention} enable joint modeling of video frames and textual inputs, potentially supporting holistic evaluation that integrates perception, semantics, and instruction adherence. However, integrating VLMs into AIGVE raises challenges, including hallucination risks, domain misalignment, and high computational cost. Future work should focus on grounding VLM outputs with structured evaluation signals, developing calibration strategies to reduce hallucinations, and designing lightweight or modular evaluation pipelines that retain reasoning capabilities while improving efficiency. Such directions may allow VLMs to move beyond superficial alignment scoring toward deeper, explainable evaluation.

\textbf{Enhancing Score Interpretability and Diagnostic Transparency}.  
A major limitation of current evaluation systems lies in their reliance on aggregated or unified scores produced by complex neural architectures. While these models often achieve strong correlations with human judgment, their opaque decision processes reduce trust, hinder debugging, and limit their usefulness for guiding model improvement. This lack of transparency is particularly problematic when evaluation results influence creative decisions, platform moderation, or revenue allocation.

Recent studies have begun exploring interpretable scoring frameworks \cite{hu2023tifa, lu2024llmscore, fu2023gptscore, ku2023viescore}, but systematic solutions remain underdeveloped. Future research should investigate multi-aspect scoring systems that provide structured breakdowns across perceptual, semantic, and temporal dimensions, coupled with natural language explanations or visual evidence grounding. Incorporating explainable AI techniques \cite{gao2023interpretabilitymachinelearningrecent, wang2024largelanguagemodelsinterpretable} and attribution mechanisms could further illuminate how evaluation signals are derived. Additionally, hybrid human–AI evaluation loops may improve interpretability by aligning machine reasoning traces with human feedback. These directions aim to transform evaluation outputs from opaque judgments into actionable diagnostics.

\textbf{Addressing Ethical Risks, Bias, and Safety Constraints}.  
Current AIGVE frameworks inherit biases and ethical risks from both training data and annotation processes. Human-labeled scores reflect subjective perspectives shaped by cultural and demographic factors \cite{Sun2023Aligning, Mittag2021Bias-Aware}, which can lead to unfair or skewed evaluations. Moreover, most existing systems lack mechanisms to assess harmful content generation, misinformation risks, or societal impact, limiting their suitability for deployment in open environments.

To address these challenges, future research should emphasize fairness-aware dataset construction and bias-aware modeling strategies. Diversifying annotation sources, applying stratified sampling \cite{Du2021Robust, Karakas2023Automated}, and integrating fairness-aware learning algorithms \cite{zliobaite2017} may reduce systematic distortions. In parallel, safety-aware evaluation modules should be embedded into AIGVE pipelines to detect harmful, misleading, or sensitive content. Privacy-preserving training and evaluation practices — including anonymization, secure data handling, and compliance with regulatory standards — are also critical for maintaining public trust. Ultimately, embedding ethical reasoning directly into evaluation frameworks represents a key step toward responsible deployment.

\textbf{Toward Unified and Scalable Evaluation Frameworks}.  
A broader limitation across the field is the fragmentation of benchmarks, metrics, and modeling approaches, which often evaluate narrow aspects of video quality without integration. This fragmentation complicates cross-study comparison and prevents the establishment of standardized evaluation protocols. Future progress will require the development of unified, modular frameworks that combine perceptual, semantic, temporal, and safety-oriented assessment under shared benchmarking standards. Collaborative community-driven benchmarking and open evaluation platforms may further support reproducibility and comparability across studies.

In summary, advancing AIGVE requires addressing fundamental limitations in multimodal reasoning, interpretability, fairness, and methodological fragmentation. By focusing on grounded integration of foundation models, transparent scoring mechanisms, ethical safeguards, and standardized evaluation infrastructures, future research can move toward more reliable and responsible evaluation systems for AI-generated video content.










%% file: content/10-conclusion.tex
\section{Conclusion}

This survey highlights the importance of AI-Generated Video Evaluation (AIGVE) as a distinct research area, focusing on aligning AI-generated videos with human perception and instructions. By reviewing existing methodologies from video quality assessment, multimodal text-visual alignment, and recent comprehensive evaluation approaches, we provide a structured overview of the current landscape. As AI-generated video technology continues to advance, there is a critical need for developing more robust evaluation frameworks that effectively capture the complexity of both spatial and temporal dimensions in video content while ensuring alignment with human needs. We hope this survey serves as a foundational resource for researchers, supporting the advancement of this evolving field.